\documentclass[acmtog, nonacm]{acmart}

\usepackage{booktabs} % For formal tables
\usepackage[utf8]{inputenc}
\usepackage{float}
\usepackage{afterpage}
\usepackage{amsmath}
\usepackage{color}
\usepackage[sectionbib]{bibunits}
\defaultbibliographystyle{plainnat}
\defaultbibliography{references}
\usepackage{algorithm}
\usepackage{multicol}
\usepackage{makecell}
\usepackage{multirow}
\usepackage{array}
\usepackage{subfig}
\usepackage{stfloats}
\usepackage{graphicx}
\usepackage{tikz}
\usepackage{enumitem}
\usepackage{fontawesome5}
\usepackage{manfnt}
\usepackage[export]{adjustbox}
\usepackage[font=small,skip=5pt]{caption}
\usepackage[noend]{algpseudocode}

\newcommand{\selthree}{S^{\,\mbox{\footnotesize \mancube}}}
\newcommand{\seltwo}{S^{\scriptsize \boxplus}}
\newcommand{\seltwoauto}{S^{\scriptsize \boxdot}}

\newcommand{\bmu}{\boldsymbol{\mu}}

\usepackage{amsfonts}

\makeatletter
\DeclareRobustCommand\onedot{\futurelet\@let@token\@onedot}
\def\@onedot{\ifx\@let@token.\else.\null\fi\xspace}

\makeatother

\usepackage{xcolor}
\definecolor{darkred}{rgb}{0.7,0.2,0.1}
\definecolor{darkgreen}{rgb}{0,0.7,0}
\definecolor{orange}{RGB}{255,127,0}
\definecolor{ourpurple}{RGB}{127,127,204}
\definecolor{ourteal}{RGB}{60,160,160}
\definecolor{palgreen}{RGB}{51,179,179}
\definecolor{magenta}{RGB}{199,21,133}

\newcommand{\ourmodel}{ArtisanGS}

\newcommand{\yes}{\checkmark}%
\AtBeginDocument{%
  }

\setcopyright{none} %{acmlicensed}
\acmISBN{978-1-4503-XXXX-X/18/06}

\acmSubmissionID{papers\_495}

\setcitestyle{square}

\begin{document}

\title{{\ourmodel}: Interactive Tools for Gaussian Splat Selection
\\with AI and Human in the Loop}

%AI-Assisted Segmentation and Completion
%of Gaussian Splat Objects with Human-in-the-Loop}

% Interactive Tools for Gaussian Splat Selection
% and Editing with AI and Human in the Loop

% CraftGS: Interactive AI-powered Segmentation and Refinement for Crafting High-Quality Gaussian Splat Objects from In-the-Wild Reconstructions
% 
% User and AI Assisted Distillation of Interactive 3D Objects
% from 3D Gaussian Splat Reconstrucitons
%
% Verbs: crafting, curating, fashioning
% Segmentation, Object Completion, Refinement
%
% Pithy Tagline: Description
\author{Clement Fuji Tsang}
\email{caenorst@hotmail.com}
\orcid{0009-0002-0998-1581}
\affiliation{%
  \institution{NVIDIA}
  \country{Canada}
}

\author{Anita Hu}
\email{anitah@nvidia.com}
\orcid{0009-0004-2275-4580}
\affiliation{%
  \institution{NVIDIA}
  \country{Canada}
}

\author{Or Perel}
\email{orr.perel@gmail.com}
\orcid{0000-0001-6478-1422}
\affiliation{%
  \institution{NVIDIA and University of Toronto}
  \country{Canada}
}

\author{Carsten Kolve}
\email{carsten.kolve@gmail.com}
\orcid{0009-0002-8601-0260}
\affiliation{%
  \institution{NVIDIA}
  \country{France}
}

\author{Maria Shugrina}
\email{shumash@gmail.com}
\orcid{0000-0002-7583-6772}
\affiliation{%
  \institution{NVIDIA and University of Toronto}
  \country{Canada}
}



\begin{abstract}

Representation in the family of 3D Gaussian Splats (3DGS) are growing into a viable alternative to traditional graphics for an expanding number 
of application, including recent techniques that
facilitate physics simulation and animation. However, extracting usable objects from in-the-wild captures remains challenging and controllable editing techniques for this representation are limited. Unlike the bulk of emerging techniques, focused on automatic solutions or high-level editing, we introduce an interactive
suite of tools centered around versatile Gaussian Splat selection and
segmentation. We propose a fast AI-driven method to propagate user-guided 2D selection masks to 3DGS selections. This technique allows for user intervention in the case of errors and is further coupled with flexible manual selection and segmentation tools. These allow a user to achieve virtually any binary segmentation of an unstructured 3DGS scene. We evaluate our toolset against the state-of-the-art for Gaussian Splat selection and demonstrate their utility for downstream applications by developing a user-guided local editing approach, leveraging
a custom Video Diffusion Model. With flexible selection tools, users
have direct control over the areas that the AI can modify. Our selection and editing tools can be used for any in-the-wild capture without additional optimization. 
  \end{abstract}

\begin{teaserfigure}
\centering
  \includegraphics[width=0.98\textwidth]{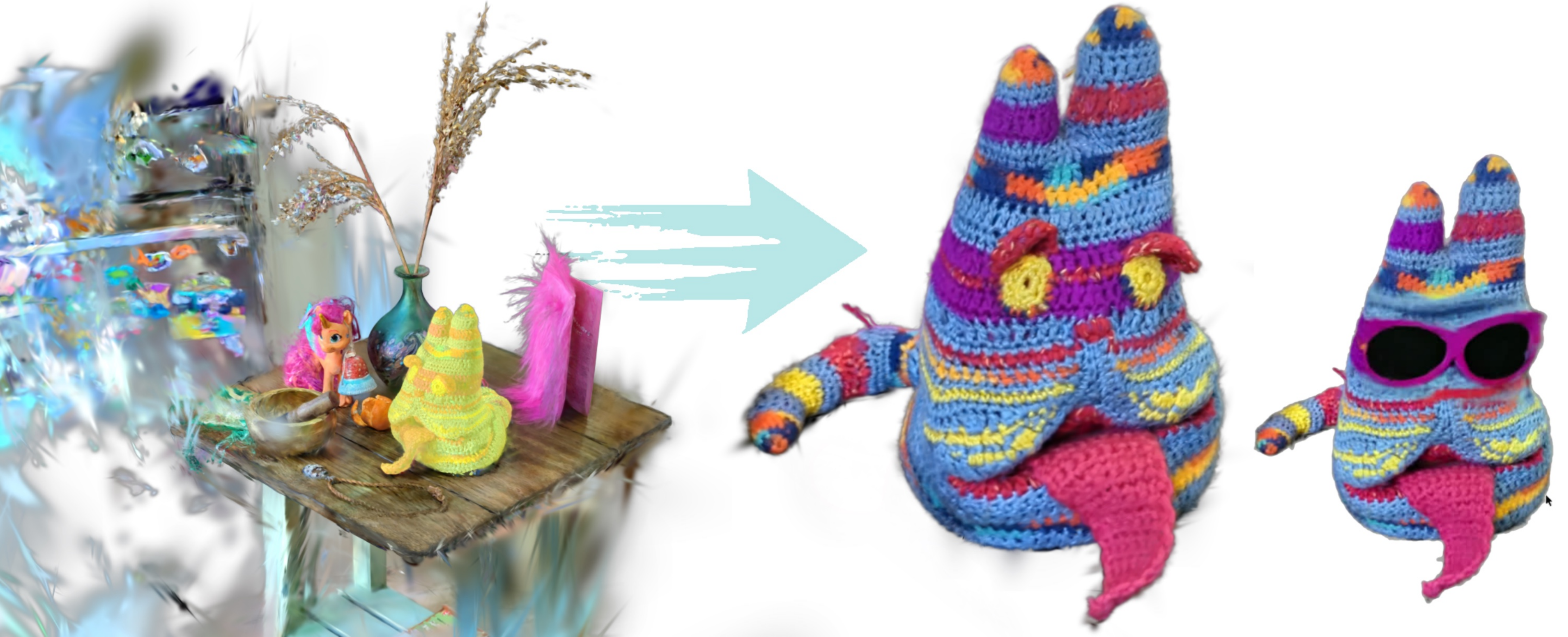}
  \caption{\textbf{From messy kitchens to interactive objects:} Our goal is to enable 3D practitioners to source objects from in-the-wild captures and work with them for emerging downstream applications, such as editing and physics simulation. To touch the tip of the iceberg on this topic, we propose a suite of interactive techniques for selection of objects and parts in 3D Gaussian Splat scenes, enabling applications like targeted editing. }
  \label{fig:teaser}
\end{teaserfigure}

\maketitle
\begin{bibunit}[ACM-Reference-Format]
    \section{Introduction}

Recent methods for multi-view capture, such as 3D Gaussian Splats (3DGS) \cite{kerbl3Dgaussians}, have reduced the barrier around capturing high-fidelity realistic 3D scenes. 
With other advances, such as representation-agnostic physics
simulation \cite{modi2024simplicits} and generative capabilities \cite{yi2023gaussiandreamer},
3DGS representation
and its variants \cite{moenne20243d,huang20242d} are on track to become a feasible alternative to mesh-based graphics for many interactive applications. For example, interactive 3D scenes 
could be authored from videos captured in the wild;
realistic simulation environments for robotics could be
created by simply recording the real world. However, a tangible
lack of tools for processing 3DGS captures into interactive
scenes, including fine segmentation and targeted editing tools, hinders these applications in practice.  

We propose {\ourmodel},
a suite of AI-powered and user-driven tools for flexible selection of 3DGS objects from unstructured in-the-wild captures (\S\ref{sec:segment}). Our selection tools are designed to work in conjunction with other applications (\S\ref{sec:applications}) needing local control like 'editing'. We enable workflows where 3DGS objects are separated from the rest of the scan, processed and then re-composed into the original 3d capture or novel environment - which in turn enables interactive applications for games and robotics. 

Working with monolithic 3DGS captures is challenging due to their unstructured nature. To enable object-based interactions, Gaussians representing specific objects must be \textbf{segmented}. While several prior works address AI-based semantic grouping and masking of Gaussians (Tb.\ref{tb:seg_methods}),
most require lengthy training and none offer strategies to correct mistakes, which makes these
methods difficult to apply in practice. In contrast, emerging commercial tools (see Supplemental) primarily target
laborious manual operations. We design versatile interactive tool that allows a user to work with minimal-input AI-assisted 3D segmentation (often just one or two clicks), diagnose and correct mistakes, and manually target specific areas. 
We develop a clean and consistent treatment of 2D and 3D selections in our toolkit design, enabling users to jump between different selection
modes. Together, these techniques
enable 3D artists to achieve nearly any desired 3D segmentation fast.

Flexible 3D selection empowers many applications (\S\ref{sec:applications}) that benefit from targeted control.  As one example, we develop local editing
of segmented 3DGS objects guided by user selections with a custom object-centric Video Diffusion Model. We also show how selection tools can facilitate guided scene orientation and enable scene setup and material assignment for applications like physics simulation, now possible directly over 3DGS objects \cite{modi2024simplicits}.

In summary, our contributions are as follows:
\begin{itemize}
    \item Interactive 3D segmentation method for 3DGS objects, powered by pre-trained 2D segmentation networks, requiring a single
    click or 2D mask for any one view (no offline scene-based optimization)
    \item Quantitative and qualitative evaluation of this selection against baselines and commercial software
    \item Technique to allow users to diagnose and correct errors in the above nearly automatic segmentation
    \item Interactive techniques allowing users to employ a 2D segmentation network or traditional 2D segmentation tools for manual clean-up of 3D segmentation
    \item Demonstration of possible applications that could be facilitated by flexible 3DGS selection
\end{itemize}
Our method can be applied to any messy in-the-wild-capture, without requiring original training views or scene-specific pre-training. We prototype an interface to show the utility of our selection and segmentation toolkit design.

% https://mirror.quantum5.ca/CTAN/fonts/fontawesome5/doc/fontawesome5.pdf
\newcommand{\methodrow}{& & & & &}
\newcommand{\inpclick}{\faHandPointer\:}
\newcommand{\inpclickfixed}{\textcolor{darkred}{\faHandPointer\:}}
\newcommand{\inppaint}{\faPaintBrush\:}
\newcommand{\inpbox}{\faBorderStyle\:}
\newcommand{\inpmask}{\faMask\:}
\newcommand{\inpmasksfixedview}{\textcolor{darkred}{\faTheaterMasks\:}}
\newcommand{\inpmasks}{\faTheaterMasks\:}
\newcommand*\inpclickone{\includegraphics[height=8pt]{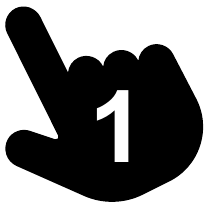}\:\:}
\newcommand{\inpscribble}{\faSignature\:}
\newcommand{\inpquery}{\faComment[regular]\:}
\renewcommand{\yes}{\textcolor{black}{yes}}
\newcommand{\na}{n/a}
\newcommand{\inp}

\begin{table}[hbp]
\centering

\begingroup
\small{
\begin{tabular}{@{}|lr||l|l|}
\Xhline{2.0pt}
\textbf{Method} & & \textbf{Features} & \textbf{Seg.}\\
 & & \textbf{training} & \textbf{time}\\
\Xhline{2.0pt}
ClickGaussian &\cite{choi2025click} & yes & 10 ms\\  %13min
OmniSeg3D&\cite{ying2024omniseg3d} & yes & not reported\\ %30-40min
GARField &\cite{kim2024garfield} & yes (bndl) & 320ms / lvl\\ %23-30min 
Gau-Grouping &\cite{ye2025gaussiangrouping} & yes (bndl) & 1.2s\\ %9min
EgoLifter&\cite{gu2025egolifter} & yes (bndl) & not reported\\ %130min
%RT-GS2&\cite{jurca2024rtgs2} & general & 37ms (2D) & no & \na & no \\
Feature3DGS &\cite{zhou2024feature} & yes & not reported\\ %6-48min
FlashSplat&\cite{shen2025flashsplat} & no & 30s\\
GaussianEditor&\cite{chen2024gaussianeditor} & no & 40s\\
iSegMan&\cite{zhao2025isegman} & yes & 4-6s\\
Seg-Wild&\cite{segwild2025} & yes & not reported\\
\Xhline{1.0pt}
\textbf{pre-prints} & & & \\
SAGA &\cite{cen2023saga} & yes & 4ms\\ %10-40min
CoSSeg &\cite{dou2024cosseggaussians} & yes & not reported\\
%Grad.-Driven&\cite{joji2024gradient} \methodrow \\  % paper too poor quality
GaussianCut&\cite{jain2024gaussiancut} & no & 50-120s\\
\Xhline{2.0pt}
\textbf{Ours} & & no & 1-5s\\
\Xhline{2.0pt}
\end{tabular}
}
\endgroup
\caption{3D Segmentation methods for 3D Gaussian Splats, prioritizing methods that take some form of user input. Many approaches require Features training or extraction for every input scene, where in some cases this step is bundled into original 3DGS training (noted as "(bndl)"). When reported, segmentation time is listed, given user input such as a target click.}
\label{tb:seg_methods}
\end{table}

\section{Related Work}\label{sec:related}

\subsection{Multi-View 3D Capture}

Many recent methods focused on optimizing a 3D representation of a scene based on multi-view
photographs. Following a rich line of work based on Neural Radiance Fields (NERFs) \cite{mildenhall2021nerf}, a recently popular 3D representation
for multi-view 3D captures is 3D Gaussian Splatting (3DGS) \cite{kerbl3Dgaussians}. While, like NERF,
vanilla 3DGS suffers from baked lighting, which becomes obvious during scene manipulation, many emerging works are
extending this representation to be more in line with physics based rendering \cite{moenne20243d,du2024gsid,gao2025relightable,liang2024gsir}.
Unlike NERF, its many variants \cite{mueller2022instant,wang2023adaptive}, and
grid-based alternatives such as \cite{fridovich2022plenoxels}, which are difficult to manipulate locally,
the explicit nature of 3DGS makes this representation family \cite{kerbl3Dgaussians,huang20242d,rong2024gstex,moenne20243d} convenient for editing and local transformations. 
This has led to an exciting line of work that uses 3DGS for physical interaction and direct manipulation
\cite{modi2024simplicits,xie2023physgaussian,zhao2024simanything,jiang2024vr}, or allows 
sophisticated editing \cite{chen2024gaussianeditor,liu2024stylegaussian}. However, a prerequisite for most
applications of this kind is segmenting individual objects or parts from the 3DGS scene, 
where existing methods fall short in flexibility.

\subsection{Segmenting 3D Gaussian Splats}\label{ssec:prev_seg}

A wide range of works address segmentation of 3DGS scenes, 
with the most notable tabulated in Tb.\ref{tb:seg_methods}.
Many prior methods devise per-scene learning techniques (Tb.\ref{tb:seg_methods}, per-scene training) to counter mask inconsistencies of 2D segmentation networks like SAM
\cite{kirillov2023segment}, resulting in per-Gaussian feature vectors.
Feature3DGS \cite{zhou2024feature} distill general 2D features, not focusing specifically on segmentation, while other approaches
of this class typically use SAM or another 2D segmentation network.
GaussianGrouping \cite{ye2025gaussiangrouping} propagates IDs using off-the-shelf tracker \cite{cheng2023tracking}, and imposes regularization losses.
ClickGaussian \cite{choi2025click}, GARField \cite{kim2024garfield}, SAGA \cite{cen2023saga}, SegWild \cite{segwild2025} and OmniSeg3D \cite{ying2024omniseg3d} all perform scale-aware or hierarchical contrastive learning
to ensure multi-view feature consistency, while EgoLifter's \cite{gu2025egolifter} contrastive learning takes special care of dynamic objects.
These approaches are effective at generating a pre-processed version of the scene that could be used for semantic reasoning and once
trained allow segmenting out 3D Gaussians based on queries such as feature proximity to one or several points. Beyond the time-consuming scene pre-processing, which takes tens of minutes to hours, depending on the method, this class of techniques is inherently limited in their flexibility. While these approaches
implement slightly different interfaces and strategies for converting features to 3D masks, mistakes, noise and biases are inevitably
pre-baked into the feature field. It is impossible for the user to deviate from the types of masks used for pre-training or
to correct mistakes beyond adding additional target feature vectors.  Many of the same limitations apply to RT-GS2 \cite{jurca2024rtgs2}, the only generalizable technique involving semantic feature learning, which trains a set of networks on a small dataset of scenes, but it is not clear if this work can generalize beyond indoor environments, and only 2D novel view segmentation results are presented. 

The approaches that generate 3D segmentation without special pre-training are more similar to our method. Like ours, most begin with a user-provided 2D mask, typically generated with a pre-trained network like SAM. Several methods require initial user clicks, project these clicks onto 3D Gaussians and track these 3D points \cite{shen2025flashsplat,chen2024gaussianeditor,hu2024sagd} or epipolar line \cite{zhao2025isegman} to generate SAM queries for other views, an approach that can break down for far away views and is too tied to click-based input, making it impossible to e.g.\ manually annotate a mask. GaussianCut \cite{jain2024gaussiancut} is the only one leveraging a video mask tracking network to propagate a single 2D mask to other views, but it likewise does not offer strategies to correct mistakes. We instead choose to use Cutie \cite{cheng2024putting} for mask tracking, which due to the unique design of its memory frames, makes our interactive segmentation amenable to user correction. Once a 2D mask is extended to multi-view masks, these are aggregated to 3D Gaussian labels. GaussianEditor \cite{chen2024gaussianeditor}, SAGD \cite{hu2024sagd} (previously SAGS \cite{hu2024sags}), iSegMan \cite{zhao2025isegman} and \cite{joji2024gradient} devise voting schemes for assigning per-Gaussian labels, while FlashSplat formulates the segmentation as an integer linear programming optimization \cite{shen2025flashsplat} and GaussianCut \cite{jain2024gaussiancut} a graph cut problem. Our solution is faster than most others (See Tb.\ref{tb:seg_methods}) and easier to extend to alternative 3DGS formulations, because we treat differentiable splat renderer as a black box component. Given limited available benchmarks, our approach offers comparative quality to other others and, unlike prior methods, allows users many strategies to correct mistakes in the initial segmentation. In addition, we discuss emerging software tools in the Supplemental Material and our video.

\begin{figure}[t!]
	\centering
	\subfloat[Controlled setting]{%
		\includegraphics[width=0.45\linewidth]{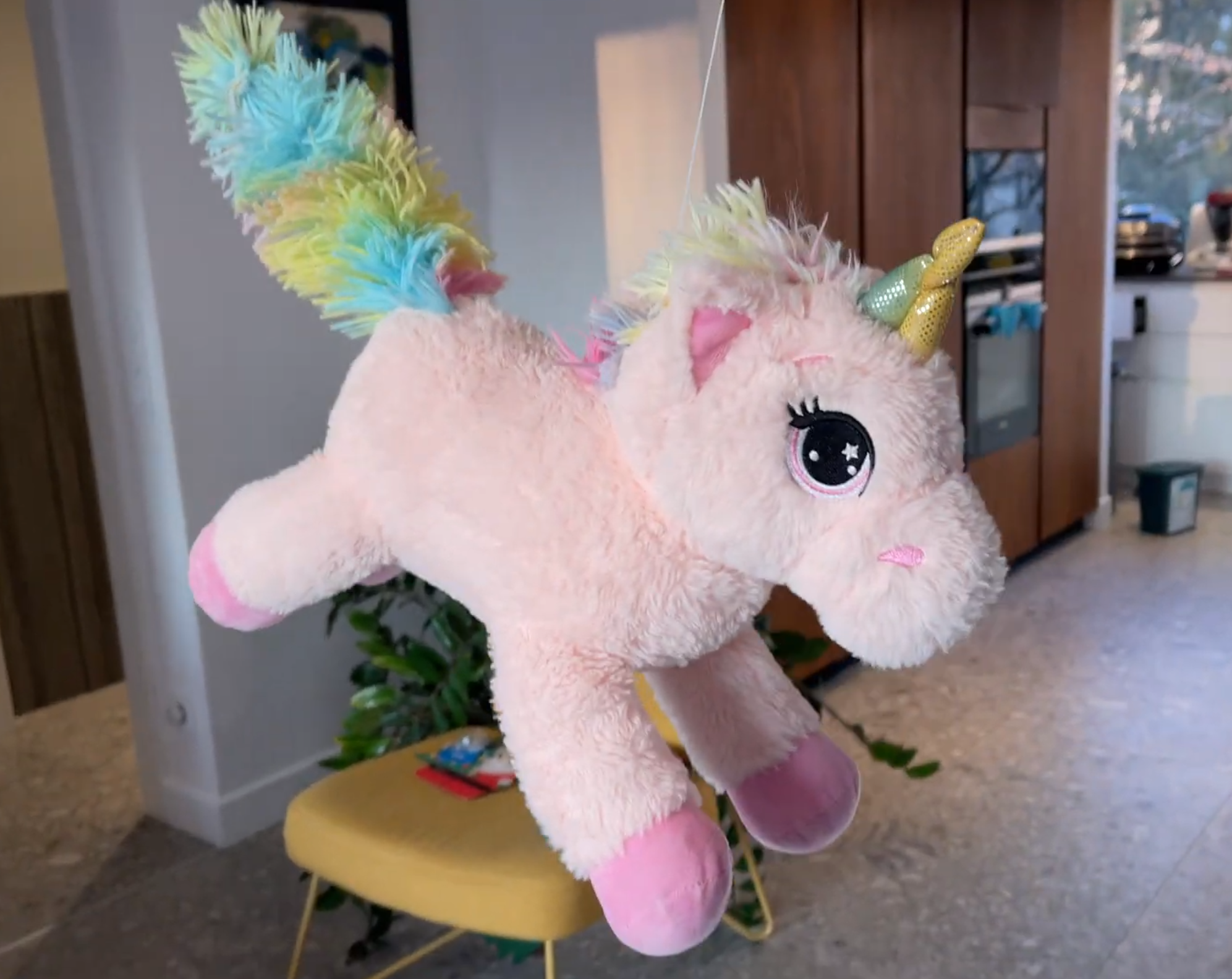}}
	\hfill
	\subfloat[Realistic setting]{%
		\includegraphics[width=0.45\linewidth]{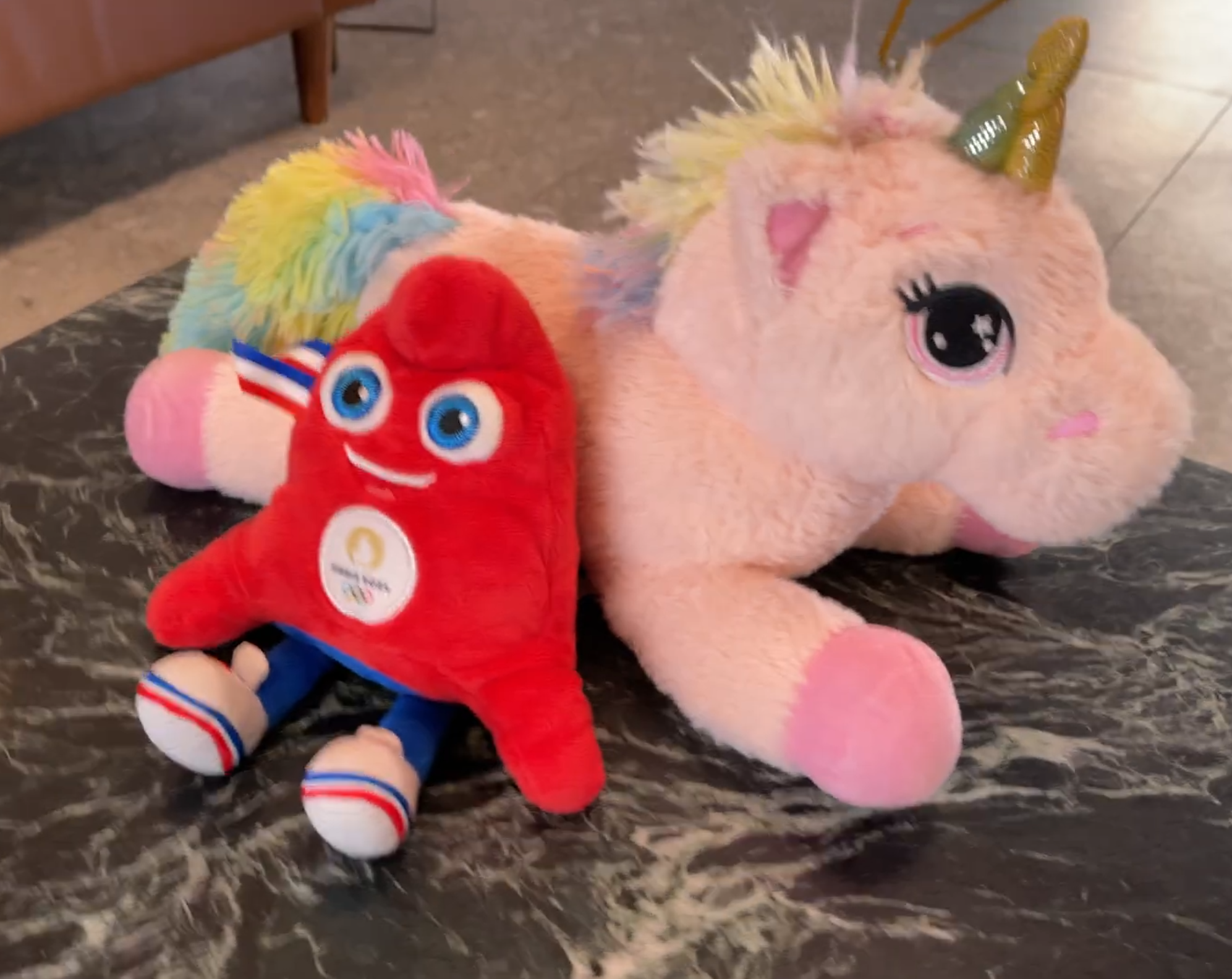}}
	\caption{\textbf{3D Capture Setups:} While segmenting objects from controlled captures, like the toy suspended by wires (a), is relatively simple with existing tools, these solutions fall short on more realistic use cases (b).}
	\label{fig:capture}
\end{figure}

\subsection{Editing Applications on Splats}\label{ssec:prev_editing}

Several approaches touch on interactive, prompt-based or style-driven editing of 3DGS scenes
\cite{chen2024gaussianeditor, yi2023gaussiandreamer, liu2023stylegaussian, igs2gs, wu2025gaussctrl, palandra2024gsedit, jiang2024vr}, but in most cases allowing users targeted control over the 
modified area is at best an afterthought. We show that the flexible selection and segmentation tools in
\ourmodel{} can enable more controllable editing applications for 3DGS. Similarly,
works targeting inpainting of 3D captured scenes \cite{chen2024mvip,lu2024view,weber2024nerfiller,barda2024instant3ditmultiviewinpaintingfast} focus on removing or hallucinating entire objects, typically assuming that the mask or target area is given. Lack of appropriate tools
makes these techniques difficult to test on in-the-wild scenes, where targeted masks are difficult
to obtain. Our goal is to enable this line of research to address more
targeted editing and completion of 3DGS scenes, to enable crafting interactive applications
from in-the-wild captures. As an illustration, we develop a prototype editing application by building on work in controllable video diffusion models \cite{hong2022cogvideo,yang2024cogvideox,xu2024camco}.

% and we demonstrate that these approaches do
% not readily extend to completing objects. 

% 2d image completion and 3d shape completion.
% then 3d conditional multi-view.

% One of our goals it to complete partially occluded 3DGS objects extracted from
% real scenes. This task is related to sparse view reconstruction, for example
% recently addressed by \cite{tang2025mvdiffusion++, xu2024grm}, however these works
% do not address the problem of occluded regions, which is the main problem with incomplete
% objects. A more extreme version of this problem is reconstructing a 3D shape from
% a single image, also popular in recent literature.
% \orp{
% Early methods leveraged score distillation signals of image diffusion models to as backbones of text-to-3d pipelines \cite{poole2022dreamfusion, lin2023magic3d}. While impressive, such methods are susceptible to issues like multiface artifacts (a.k.a the Janus problem).
% To handle view consistency, various approaches have been suggested, including fine tuning on 3D data \cite{liu2023zero1to3}, multiview-diffusion \cite{shi2023zero123plus, liu2023syncdreamer, li2024era3d, long2023wonder3d}, video diffusion models \cite{voleti2024sv3d, yang2024hi3d, chen2024v3d, melaskyriazi2024im3d}, neural SDF based parameterizations \cite{liu2023one2345, liu2023one2345++}, and direct reconstruction of meshes \cite{boss2024sf3d, zheng2024mvd}.
% }

% We incorporate advances from latest work on video diffusion models, including epipolar attention \cite{xu2024camco}.

% \cite{shi2023MVDream}

\section{Preliminaries}\label{sec:overview}

\subsection{Motivation}\label{ssec:motivation}

Multi-view captures like NERFs and 3D Gaussian Splats (3DGS) have fascinated practitioners with their ability to effortlessly reconstruct highly realistic 3D environments. However, for practical applications, these environments remained largely static, confined to novel view synthesis. Recent advances offering practical solutions for dynamic effects on 3DGS scenes, including physics simulation \cite{modi2024simplicits, xie2023physgaussian} and relighting \cite{gao2025relightable}, are promising to bring 3DGS into dynamic applications. While individual high-quality 3DGS objects can be reconstructed from highly controlled capture setups (e.g.\ suspended by wires in Fig.\ref{fig:capture}a), this severely limits ability to capture more realistic scenes (Fig.\ref{fig:capture}b) or work with videos and splat models available online. Extracting usable objects from such in-the-wild 3DGS scenes is prohibitively difficult with the current state of tools, a gap that we address in this work.

\subsection{Method Overview}
Imagine working with a monolithic capture of a cluttered environment, like a play room. To construct an interactive environment, 
this scene must first be broken apart into individual objects. This is the core task addressed by our method in \S\ref{sec:segment}, showing why fully automatic solutions do not work perfectly, and how the user could more directly
control the final output. Flexible selection forms the necessary backbone for many applications.

Once segmented, the object or a scene might
have arbitrary orientation, making everything, starting with camera
control, more challenging. Occlusions within the scene will also inevitably cause parts of the captured objects to be missing. Unless perfectly choreographed, source views will also contain varying level of detail for different view angles,
resulting in areas of degraded quality around the object, which may need local targeted editing and refinement. In \S\ref{sec:applications}, we show
how our selection and segmentation tools can feed into such applications for 3DGS object processing, including
orientation \S\ref{sec:orient} and an early prototype of local editing \S\ref{sec:refine}. 

The clean and consistent design of 2D and 3D selection in our toolkit combines automatic AI-driven techniques with user input and could be integrated into future end-user applications for crafting interactive 3DGS scenes. As a sample application, we show physics simulation with Simplicits \cite{modi2024simplicits} on scenes processed with our technique (\S\ref{ssec:simplicits}).

%We present a collection of tools to enable end users to iteratively achieve perfect object segmentation within 3DGS scene (\S\ref{sec:segment}), to orient 3DGS scenes or objects based on custom criteria (\S\ref{sec:orient}) and to guide captured object completion and refinement (\S\ref{sec:refine}). Both segmentation and orientation leverage AI to make the task viable, but allow many opportunities for user intervention and iteration when needed.  

\subsection{Definitions}\label{ssec:definitions}

We start with a pre-trained 3DGS scene \cite{kerbl3Dgaussians}, containing a set $\mathcal{G}$ of $n$ individual 3D Gaussians $G_i$, each with a position $\bmu_i$, covariance $\Sigma_i$, view-dependent color and opacity $\alpha_i$. We assume the existence of optimized differentiable rendering kernels for $\mathcal{G}$, including $\mathtt{render}(\mathcal{G}, v)$ producing RGBA rendering of the 3DGS scene from camera view $v$, $\mathtt{depth}(\mathcal{G}, v)$ producing camera-space depth rendered from $v$, $\mathtt{features}(\mathcal{G}, v, F)$ rendering any custom per-Gaussian feature vectors $F := \{F_0...F_n\}$, $\mathtt{viz}(\mathcal{G}, v)$ producing a binary mask specifying the Gaussians visible from $v$ and $\mathtt{first}\_\mathtt{hits}(\mathcal{G}, v)$ outputting the id of the first hit Gaussian per pixel. Our method does not assume anything beyond the existence of these functions and so is in principle applicable to alternative 3DGS variants such as \cite{huang20242d,moenne20243d}, but we have run all our experiments on the original formulation \cite{kerbl3Dgaussians}, trivial to extend to $\mathtt{depth}(\mathcal{G}, v)$ and $\mathtt{features}(\mathcal{G}, v, F)$ by applying the original $\mathtt{render}(\mathcal{G}, v)$ function to other per-Gaussian features.

Typically, $\mathcal{G}$ has high-quality appearance from
camera views that do not stray too far from the training views $\bar{V} := \{\bar{v}_0...\bar{v}_n\}$ used for optimization, but may appear foggy or abstract from views that are not well-represented (Fig.\ref{fig:teaser}), which can affect the result of AI models trained on real images. While we do not assume the knowledge of $\bar{V}$, our algorithm has the option of using them if available. 

%The goal of our interactive segmentation is to enable a user to interactively select individual Gaussian subsets
%splat objects $\mathcal{O}_i$ from $\mathcal{G}$, such that each object $\mathcal{O}_i$ contains imagery that follows user's intent (for example, whole object, or a very specific part) and has good appearance from any view angle around the object, thus enabling its use in interactive applications where the object can be moved and simulated. 

%In addition, we require the presence of a fast shader $\mathtt{first\_hits}(\mathcal{G}, C, \tau_{\alpha})$, that for a given camera $C$ outputs the ID of the first hit Gaussian per pixel, subject to a minimum opacity value of $\tau_{\alpha}$. \todo{Check correctness, add a note on implementation -- did we add a custom shader?}

    \begin{figure*}[ht!]
	\centering
		\includegraphics[width=0.99\linewidth]{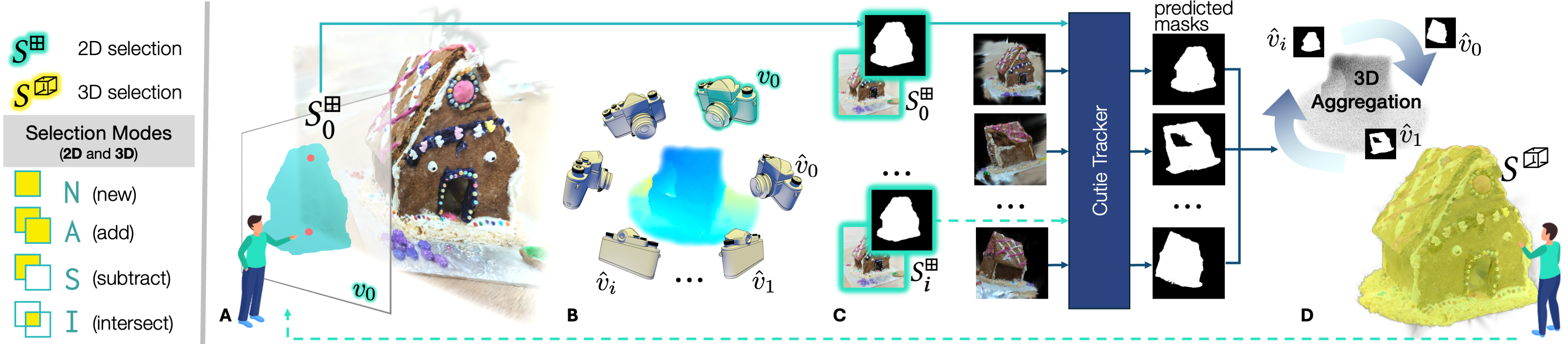}
	\caption{\textbf{Auto-Tracked Segmentation with Corrections}: We propose automatic way to project 2D user masks $\seltwo_i$ to 3D selection $\selthree$ over 3D Gaussians, while allowing users to correct the outcome (\S\ref{ssec:seg_auto}). \textbf{Left:} notation and selection modes supported in our design (\S\ref{sec:segment}).}
	\label{fig:segment_auto}
\end{figure*}

\section{Interactive Segmentation}\label{sec:segment}

\begin{figure}[ht!]
	\centering
	\subfloat[Frustum projection]{%
		\includegraphics[width=0.57\linewidth]{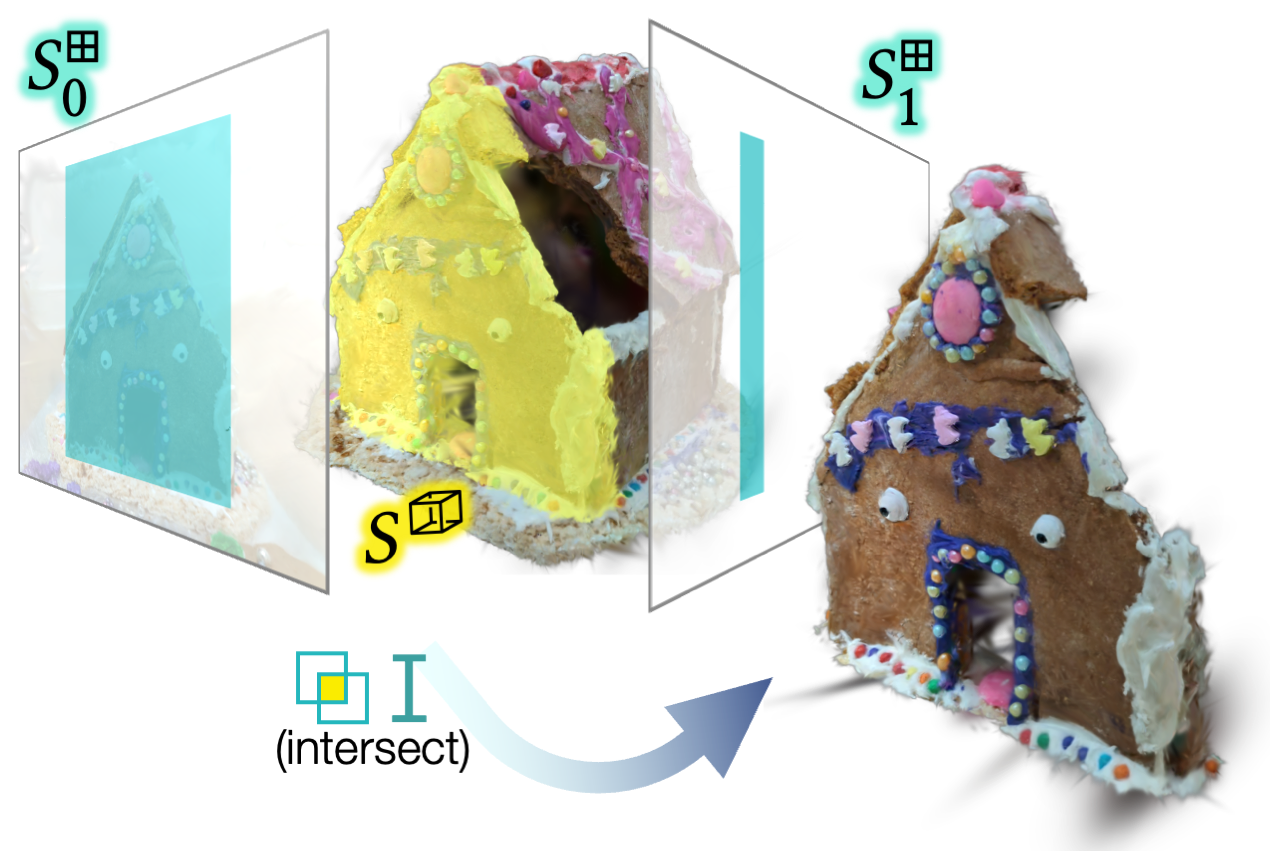}}
	\subfloat[Depth projection]{%
		\includegraphics[width=0.40\linewidth]{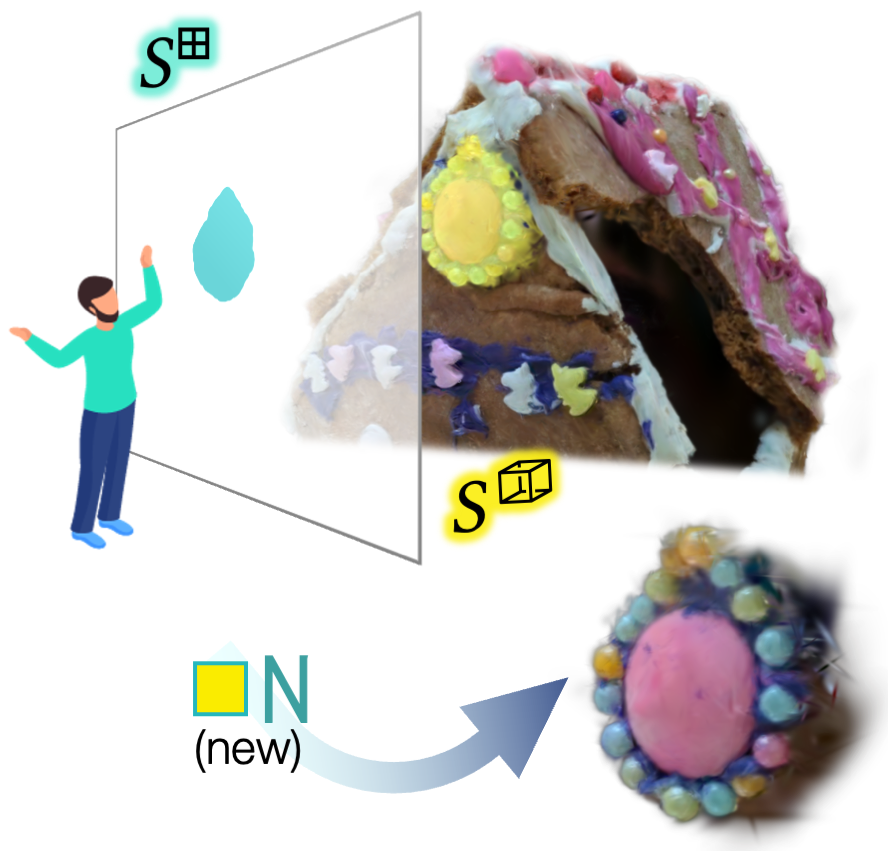}}
	\caption{\textbf{Manual Projection}(\S\ref{ssec:seg_manual}) of 2D masks $\seltwo_i$ to 3D, combined with different selection modes (\S\ref{ssec:seg_modes}), allow flexible manual selection.}
	\label{fig:segment_manual}
\end{figure}

We will now present an interactive toolkit for 3DGS selection and segmentation.
We define the segmentation problem as selecting a subset of Gaussians
$\mathcal{S} \in \mathcal{G}$ according to the user intent. 
Any implementation of our proposed toolkit will allow users control of the
camera view $v_c$ and will keep track of the 2D segmentation mask $\seltwo$ (active for $v_c$) and of the
current 3D segmentation mask $\selthree$, containing a binary
value for every element of $\mathcal{G}$. We will use $\mathcal{S}$ to refer to the subset of Gaussians currently in $\selthree$. Different color is used to denote 2D and 3D selections in our diagrams and in our UI to make them easier to distinguish (See Fig.\ref{fig:segment_auto}).
%\todo{define frustrum here}

Given preliminary requirements (\S\ref{ssec:seg_modes}, \S\ref{ssec:seg_2d}), we propose 
combining multiple ways to project 2D masks to $\selthree$. First, we show simple yet powerful manual projection
modes (\S\ref{ssec:seg_manual}), and then detail our automatic mask tracking and segmentation method in \S\ref{ssec:seg_auto}.

\subsection{Selection Modes}\label{ssec:seg_modes}

\newcommand{\selectmode}[1]{\texttt{\color{ourteal}{#1}}}
Established image selection tools in software like Photoshop support the following major modes:
 new (\selectmode{N}) - replace current selection, add to (\selectmode{A}), subtract from (\selectmode{S}) and intersect with (\selectmode{I})
 the current selection. To provide consistent experience,
we support all of these modes both for 2D selection and 3D selection (Fig.\ref{fig:segment_auto}, left). In both cases, the modes
correspond to boolean operations on the active image mask $\seltwo$ or binary per-Gaussian mask $\selthree$. In our implementation, \selectmode{N} is
the default mode, and others are activated through \texttt{Ctrl}, \texttt{Alt} and \texttt{Shift} modifiers, but many other UI 
designs are possible.

%Notations:  \selectmode{N} \selectmode{A} \selectmode{S} \selectmode{I} $\seltwo$ $\selthree$ $\seltwo_0$ $\seltwo_1$ $\seltwo_i$ $v_c$

\subsection{Required 2D Mask Capabilities}\label{ssec:seg_2d}

Because users can only see a 2D rendering of $G$, 2D selection is a necessary prerequisite for any 3D selection
interface. While any 2D image segmentation techniques could be used in combination with our toolkit, it is critical to allow users
a variety of 2D masking tools to bootstrap interactive 3D segmentation below. Our particular implementation, 
like others before it, allows users to generate a mask $\seltwo$ for any view $v_c$ 
from one or more positive and negative clicks using SAM \cite{kirillov2023segment}, but
other models could be used instead. We also allow manually painting 2D areas and drawing 2D bounding boxes, while supporting the selection modes 
 in \S\ref{ssec:seg_modes}. 

\subsection{Manual 2D to 3D Projection Modes}\label{ssec:seg_manual}

In addition to automatic aggregation (\S\ref{ssec:seg_auto} to follow), our design supports two manual modes of projecting user's
2D selection to 3D. The \emph{frustum projection} mode selects all Gaussians, for which the mean projects into the current 
mask $\seltwo_c$ of the current view $v_c$. This effectively selects all the Gaussians falling into the sweep of $\seltwo_c$ through
the camera frustum. Coupled with the selection modes in \S\ref{ssec:seg_modes}, frustum projection 
is surprisingly effective for some use cases. In Fig.\ref{fig:segment_manual}a, projection of $\seltwo_0$ followed by projection of $\seltwo_1$
with mode \selectmode{I} (intersect), effectively selects the facade of the gingerbread house.

In some cases, selecting surface layer is more desirable. In the \emph{depth projection} mode, we select all the Gaussians that 
are in the frustum projection of $\seltwo$ \emph{and also} lie within a threshold of the rendered depth at their location. This mode
allows picking up surface detail, such as the wreath in Fig.\ref{fig:segment_manual}b.

\subsection{Auto-Tracked Segmentation with Corrections}\label{ssec:seg_auto}

We propose a fast automatic way to convert one or more user-provided 2D masks $\seltwo_0...\seltwo_k$ corresponding to
views $v_0...v_k$ to a 3D Gaussian mask $\selthree$ (Fig.\ref{fig:segment_auto}). Like other training-free approaches for this task \cite{shen2025flashsplat,chen2024gaussianeditor,hu2024sagd,jain2024gaussiancut}, we generate multi-view masks (\S\ref{ssec:seg_auto:track}) for a dense set of views (\S\ref{ssec:seg_auto:views}) and then aggregate these masks
into a 3D mask (\S\ref{ssec:seg_auto:agg}). Uniquely, our approach allows users visibility into the method and ability to correct mistakes (\S\ref{ssec:seg_auto:correct}).

\subsubsection{Selecting Dense Views}\label{ssec:seg_auto:views}

We support multiple ways to sample dense views around the target object. If the training views $\bar{V}$ are available, 
all or a subset can be used. Training views are guaranteed to show the scene from a well-optimized angle, but may be very zoomed out
for large scenes. In addition, requiring
training views would also constrain the types of captures that could be segmented. Therefore, we devise an alternative method, using
the point cloud obtained by lifting $\mathtt{first}\_\mathtt{hits}(\mathcal{G}, v_0$ from the masked area $\seltwo_0$. We turn the camera around the center of this point cloud, doing a full circle trajectory on the plane defined by the camera up axis, and use the corresponding user view $v_0$ for up direction and distance
from the look at point. We ablate the choice of training or camera turnaround views in our results section. Note that the number of views used for tracking ($m$),
is an important hyperparameter for our method's speed, but we found it to work robustly when sampling about 50 views (See \S\ref{sec:ablations}).

%We select the views on which to predict the 2D mask by subsampling from the training views, from $\bar{V} := \{\bar{v}_0...\bar{v}_n\}$ to $\dot{V} = \{\dot{v}_0...\dot{v}_m\}$ where $m$ is evenly spaced sampled from the training views to reduce the number of views to process. 
%Subsampling from training view provides best results in the scene from LLFF and LERF dataset as the training views are usually of better quality, however it depend on the training views and so may poorly suited for large scale scene with long video where the object select might be out of frame for most of the frames. In this case we a proposing a more stable sampling method, where we generate views turning around the reference mask, we define a point cloud generated by finding the first hits (with a minimum alpha threshold value) on the masked pixels. We turn the camera around the center of this point cloud, doing a full circle trajectory on the plane defined by the camera up axis.
%\todo{Clement - just added back}

\subsubsection{Obtaining Multi-View Masks}\label{ssec:seg_auto:track}
Similarly to ours, many training-free prior methods tackling 3DGS segmentation 
start with a mask $\seltwo_0$ and then use it to obtain $m$ multi-view masks $\seltwoauto_0...\seltwoauto_m$ for the dense views $\dot{v}_0...\dot{v}_m$. Most of these techniques \cite{shen2025flashsplat,chen2024gaussianeditor,hu2024sagd} track depth-projected query points from the first mask to generate SAM queries for other views. This approach can cause errors for more extreme views where 3D query points are not visible and leaves the result at the mercy of a specific click-based network like SAM \cite{kirillov2023segment}. In contrast, we rely on a robust mask tracking network Cutie \cite{cheng2024putting}. This design decision results in more robust tracking, and supports our goal of allowing users to add more guidance when needed.

The network architecture and inference pipeline of Cutie accepts object-level conditioning through one or more reference frames, injected at appropriate points in the video sequence due to memory constraints (See user masks $\seltwo_i$ in Fig.\ref{fig:segment_auto}C). To facilitate attending to appropriate annotations, we shift the target turnaround views $\dot{v}_0...\dot{v}_m$ to begin with the optimal view angle, obtaining shifted sequence $\hat{v}_0...\hat{v}_m$ (Fig.\ref{fig:segment_auto}B). 
We select the first view $\hat{v}_0$ to be $\dot{V}^*(v_0)$, defined as the target view $\dot{v}_i$ that is most similar to the annotated view $v_0$ according to the following criterion:
\begin{equation}
    \dot{V}^*(v) := \underset{\dot{v}_j}{\mathrm{argmin}}\; J(\mathtt{viz}(\mathcal{G}, v), \mathtt{viz}(\mathcal{G}, \dot{v}_j)
\end{equation}
where $\mathtt{viz}$ is the visibility mask defined in \S\ref{ssec:definitions}, and $J$ is the Jaccard index. Thus, Cutie is injected with the user mask $\seltwo_0$ and corresponding $\mathtt{render}(\mathcal{G}, v_0)$ right before predicting the mask for the most similar view $\hat{v}_0$. Given multiple user masks $\seltwo_i$ for the views $v_i$, each $\seltwo_i$ and $\mathtt{render}(\mathcal{G}, v_i)$ is injected as a memory frame prior to predicting the corresponding most similar target view $\dot{V}^*(v_i)$. This approach allows users to annotate any number of frames for any view angle, without being constrained to pre-selected views. 

\subsubsection{Aggregating to 3D}\label{ssec:seg_auto:agg}
Given user masks $\seltwo_0...\seltwo_k$ for views $v_0...v_k$, and automatically tracked masks $\seltwoauto_0...\seltwoauto_m$ for dense views $\dot{v}_0...\dot{v}_m$, we now aggregate them into a binary mask $\selthree$ over the Gaussians (Fig.\ref{fig:segment_auto}D). Instead of devising a custom voting scheme like \cite{chen2024gaussianeditor,hu2024sagd} specific to a particular 3DGS variant, or formulating a linear programming \cite{shen2025flashsplat} or graph cut problem \cite{jain2024gaussiancut}, which has additional complexity and overhead, we simply leverage the fast differentiable 3DGS renderer to optimize the mask assignment. Specifically, we run a single loop over the views $\dot{v}_0...\dot{v}_m,v_0...v_k$, optimizing a one channel feature $M$ for each Gaussian with an L2 image loss between the 2D masks and $\mathtt{features}(\mathcal{G}, v, M)$. We found it effective to simply set the binary $\selthree$ to $M > 0.5$, a setting we use for all demos and experiments. The "black-box" treatment of 3DGS renderer makes this aggregation applicable to other point splatting formulations like \cite{moenne20243d,huang20242d}.

\subsubsection{User Corrections}\label{ssec:seg_auto:correct}
Crucially, unlike prior
methods, we also allow users to diagnose the cause of error in the 3d aggregation. Using our UI, users can browse automatic
masks generated by Cutie and add more masks for any additional view via SAM annotation or manual mask painting. The annotated views are
inserted into the Cutie inference based on their proximity with the target views for which automatic masks are being generated (\S\ref{ssec:seg_auto:track}), resulting in a more robust turnaround performance and better aggregation.

\subsubsection{Performance Improvements}\label{ssec:seg_auto:opt}
The performance of any method leveraging 2D masks for 3D aggregation is sensitive to occlusions around the object. To
improve performance of our method in cluttered scenes, we give users the option to mark a mask as containing no occlusions. When such masks are
provided (default setting), we first pre-segment the scene using intersecting frustum projections (\S\ref{ssec:seg_manual}) of these masks and
perform tracking (\S\ref{ssec:seg_auto:track}) and aggregation (\S\ref{ssec:seg_auto:agg}) only on this segment. This optimization not
only improves robustness to occlusions, but also the speed of the aggregation due to faster rendering.

\section{Results and Evaluation}\label{sec:eval}

In this section, we evaluate our auto-tracked segmentation (\S\ref{ssec:seg_auto})
quantitatively (\S\ref{ssec:eval:quant}), show qualitative comparisons with related selection and segmentation tools (\S\ref{ssec:eval:qual})
and present ablations (\S\ref{sec:ablations})
See the following section \S\ref{sec:applications} for applications.
\begin{figure}[ht!]
	\centering
	\subfloat[NVOS scribbles vs. our alternative SAM queries.\label{fig:datasets:nvos_scribble}]{%
		\includegraphics[width=0.75\linewidth]{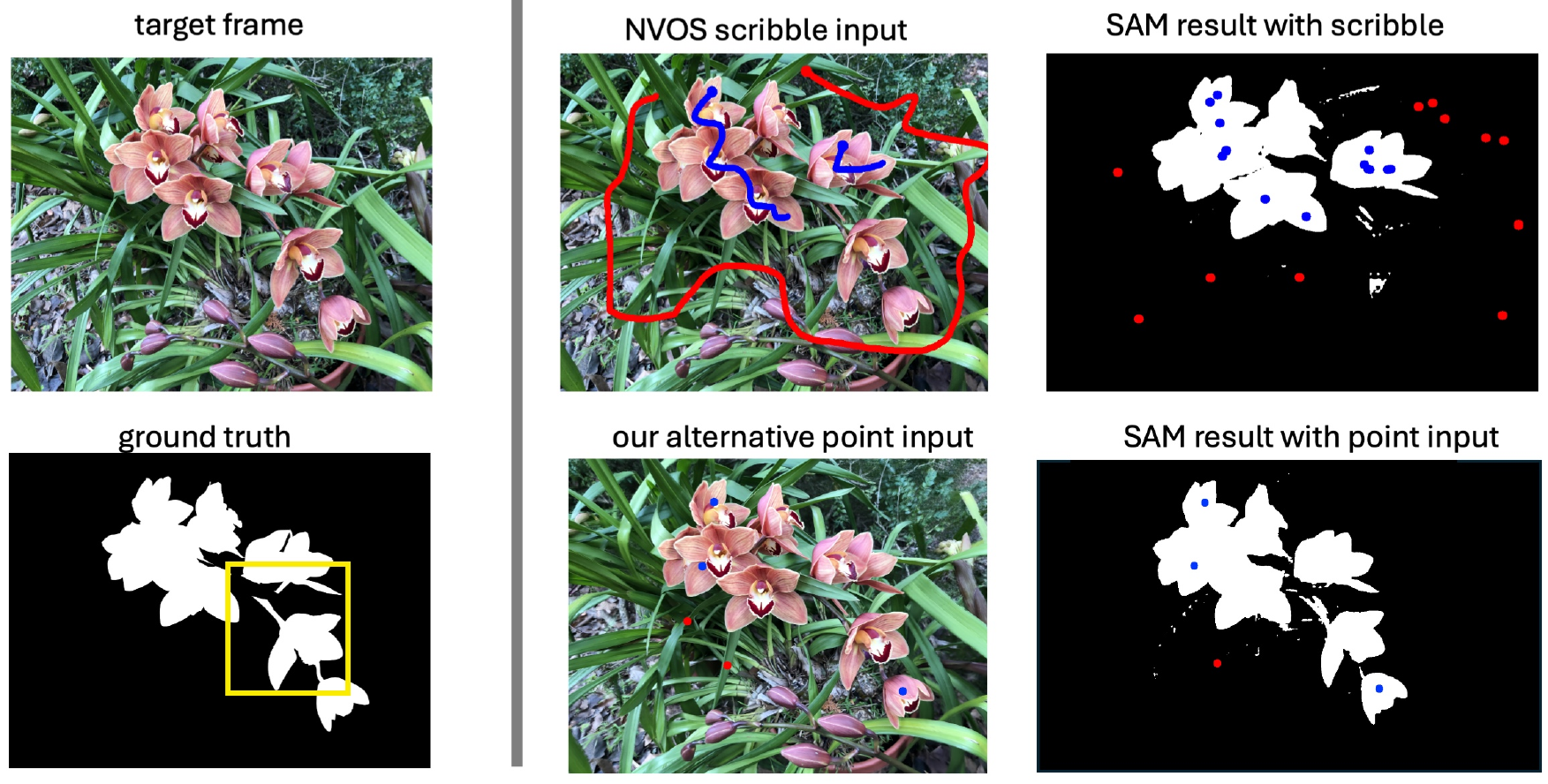}}
	\subfloat[NVOS occl.\label{fig:datasets:nvos_crop}]{%
		\includegraphics[width=0.24\linewidth]{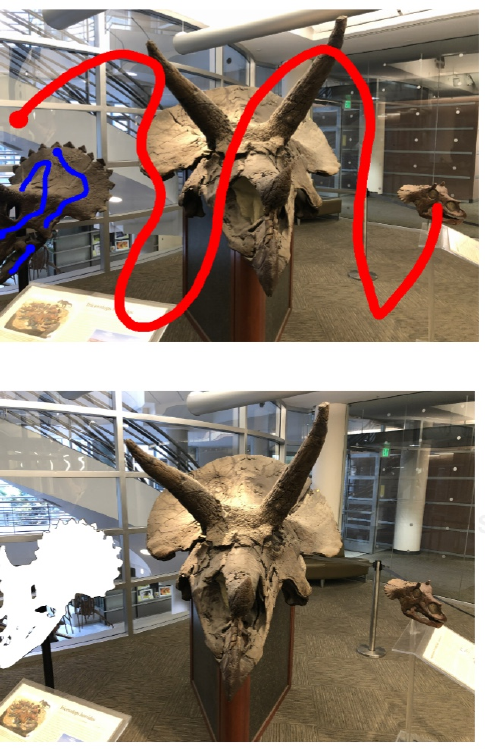}}\\
        \subfloat[Large inaccuracies in automatic masks of LERF-mask\label{fig:datasets:lerf_mask}]{%
		\includegraphics[width=0.98\linewidth]{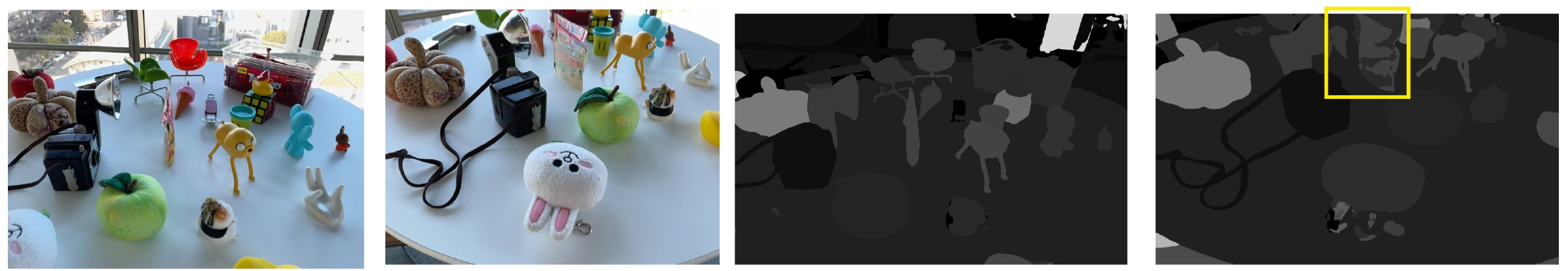}}
        %\subfloat[Manually segmented 3D Ground Truth for LERF figurines scene\label{fig:datasets:figurines_gt}]{%
		%\includegraphics[width=0.55\linewidth]{img/lerf_gt.pdf}}\\
	\caption{\textbf{Evaluation Datasets}: Annotations on both NVOS (a) and LERF-Mask (b) have inaccuracies. We suggest alternative inputs for NVOS (a)}
	\label{fig:datasets}
\end{figure}

\subsection{Quantitative Evaluation}\label{ssec:eval:quant}

\subsubsection{Baselines}
We compare our segmentation against FlashSplat\cite{shen2025flashsplat} and GaussianCut\cite{jain2024gaussiancut}, which, similarly to our method,
require no pre-processing of the input scene (See segmentation methods in Tb.\ref{tb:seg_methods}). In addition, we compare against SAGA\cite{cen2023saga}, GaussianGrouping \cite{ye2025gaussiangrouping}, iSegMan\cite{zhao2025isegman} and
OmniSeg3D\cite{ying2024omniseg3d}, competitive methods that need scene pre-processing in order to allow segmentation. Note that
our method was developed in 2024 and we may be missing some more recent techniques in this evaluation.

\subsubsection{Datasets and Metrics}\label{ssec:expsetup}
There is no robust and diverse benchmark for the segmentation task that we target. Commonly used 
NVOS \cite{ren-cvpr2022-nvos} dataset is very small, and LERF-Mask \cite{ye2025gaussiangrouping} 
contains noisy auto-labels (e.g. see noisy auto-labels in Fig.\ref{fig:datasets:lerf_mask}).
Because most baselines use this dataset, we report results on the smaller, more accurate NVOS dataset \cite{ren-cvpr2022-nvos}, consisting of a small number of front-facing scenes from LLFF \cite{mildenhall2019llff} with target scribbles in one frame and a single ground truth mask for another frame. The input
scribbles in NVOS are an imperfect representation of user intent, and fail with modern segmentation methods that rely on clicks, such as SAM \cite{kirillov2023segment} (See Fig.\ref{fig:datasets}, where scribbles are not placed on all the target flowers, confusing SAM). While many of the baselines we compare with
also rely on input clicks, the codebases and papers of FlashSplat\cite{shen2025flashsplat}, SAGA\cite{cen2023saga}, GaussianCut\cite{jain2024gaussiancut} and OmniSeg3D\cite{ying2024omniseg3d} do not include the exact logic for sampling points from NVOS scribbles or the number of points. However, these 
details have a pronounced effect on click-based methods, like SAM, and make reported numbers from these related works that are using random sampling from the scribbles unreliable. \footnote{To quantify the effect of point sampling, we ran SAM segmentation on one of the NVOS images 20 times, using randomized 3 to 10 positive and 1 to 4 negative clicks. The resulting masks have an average pairwise IoU of 68.3\%, and it can be as small as 3.5\% between runs, showing the overwhelming effect of point choices on performance.} Nonetheless, we include a comparison for completeness, and report click sampling logic here. Because our method starts with a user-annotated mask, failing due to bad SAM initialization would not meaningfully evaluate our method. Instead of NVOS scribbles, we provide a small number (1 to 6) of positive point inputs for the starting frame, and use the same ground truth frame to evaluate. For all baselines, we report original NVOS results, with
point click logic tuned for their method, and for OmniSeg3D \cite{ying2024omniseg3d}, the best available method with a released codebase, we additionally show performance on our input point clicks. We use standard metrics of pixel classification accuracy (Acc) and foreground intersection-over-union (IoU). At best, NVOS provides a very noisy estimate of performance due to its tiny size and carefully choreographed front-facing scenes, but we include it as the standard benchmark reported in literature, and later focus on qualitative differences that make our method easy to apply in practice (\S\ref{ssec:eval:qual}).

\subsubsection{Results}

Results in Tb.\ref{tb:seg_eval}, showing competitive performance of our method against baselines, while also being faster, and requiring no scene pre-processing (See Tb.\ref{tb:seg_methods} for prior method properties and speed).
Because point sampling results in noisy results (see above), we did not reevaluate the baselines under the slightly different setting of
click inputs rather than scribbles. However, for the competitive OmniSeg3D, our point samples result in severely degraded performance, suggesting
that their technique requires a lot more points to work accurately, and suggesting that custom point sampling logic of prior works is likely designed to
their advantage. In all cases, our technique outperforms others when pre-segmentation (\S\ref{ssec:seg_auto:opt} is turned off, and 
degraded performance is due to a single example ("horns left") where the annotated frame does not include the whole target object (Fig.\ref{fig:datasets:nvos_crop}). In practice, when using our method, users have a choice to enable the pre-segmentation
optimization. We next provide qualitative evaluation, showing that our technique may be easier to apply in
practice, allowing user-driven segmentation with corrections.

\begin{table}[t!]
\centering

\begingroup
\small{
\begin{tabular}{@{}|l||ll|ll|@{}}
\Xhline{2.0pt}
% & \multicolumn{2}{c|}{NVOS}  \\
\textbf{Method} & \textbf{mIoU}{$\uparrow$} & \textbf{Acc}{$\uparrow$}  \\
\Xhline{2.0pt}
FlashSplat & 91.8 & 98.6 \\
GaussianCut & 92.5 & 98.4 \\
% SAGD & 72.1 & 91.7 \\
Gaussian Grouping & 90.6 & 98.2 \\
% SA3D & 90.3 & 98.2 \\
SAGA & 90.9 & 98.3 \\
iSegMan & 92.0 & 98.4 \\
OmniSeg3D & 91.7 & 98.4 \\
OmniSeg3D (with our points) & 78.5 & 96.4 \\
\Xhline{2.0pt}
Ours (with pre-segment) & 82.4 & 98.1\\
Ours (without pre-segment) & \textbf{94.1} & \textbf{98.8}\\
\Xhline{2.0pt}
\end{tabular}
}
\endgroup
\caption{Segmentation Eval on NVOS.}
\label{tb:seg_eval}

\end{table}

\subsection{Qualitative Results}\label{ssec:eval:qual}

\begin{figure*}[h!tbp]
	\centering
	\subfloat[Segmentation pipeline.\label{fig:seg_gallery:meow}]{%
		\includegraphics[width=0.98\linewidth]{img/gallery_seg_meow.pdf}}\\
        \subfloat[Working with modes.\label{fig:seg_gallery:modes}]{%
		\includegraphics[width=0.98\linewidth]{img/gallery_seg_modes.pdf}}\\
        \subfloat[Depth projection.\label{fig:seg_gallery:depth}]{%
		\includegraphics[width=0.37\linewidth]{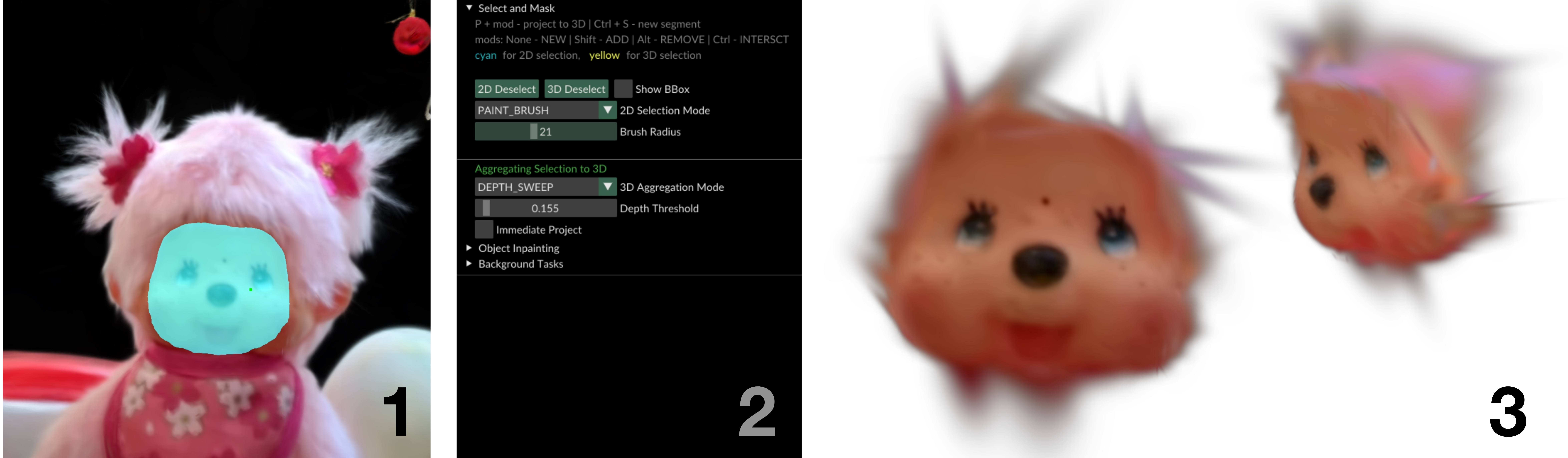}}
        \subfloat[Comparison with GaussianEditor.\label{fig:seg_gallery:gseditor}]{%
		\includegraphics[width=0.37\linewidth]{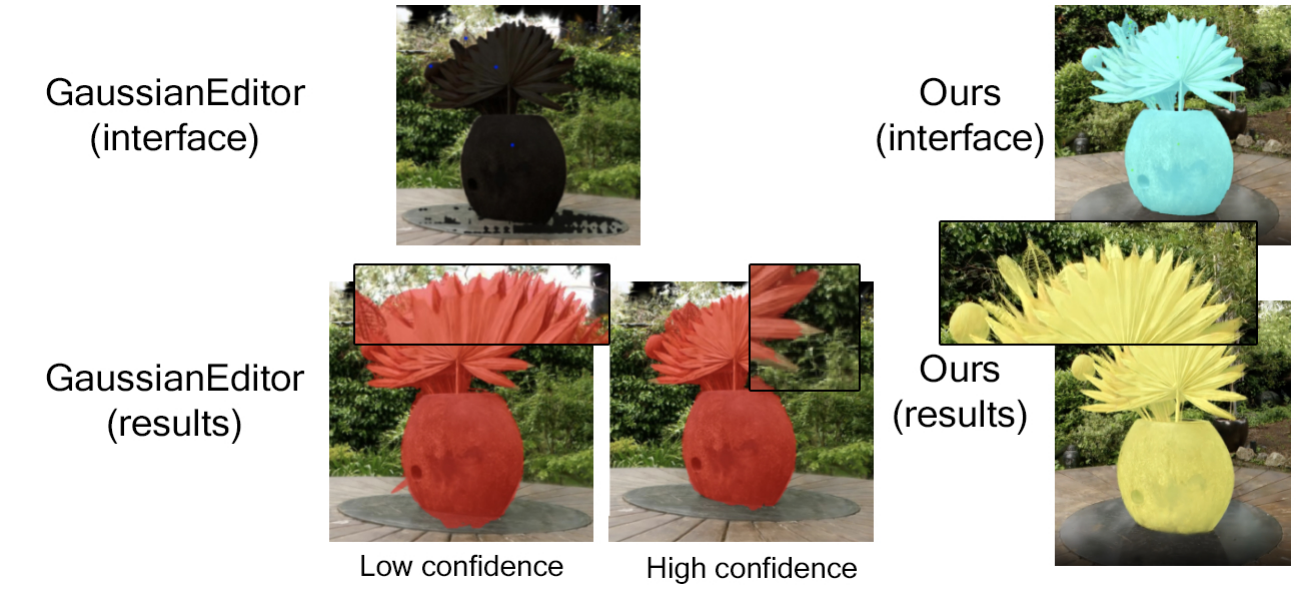}}\\
        \subfloat[Comparison with supersplat.\label{fig:seg_gallery:supersplat}]{%
		\includegraphics[width=0.52\linewidth]{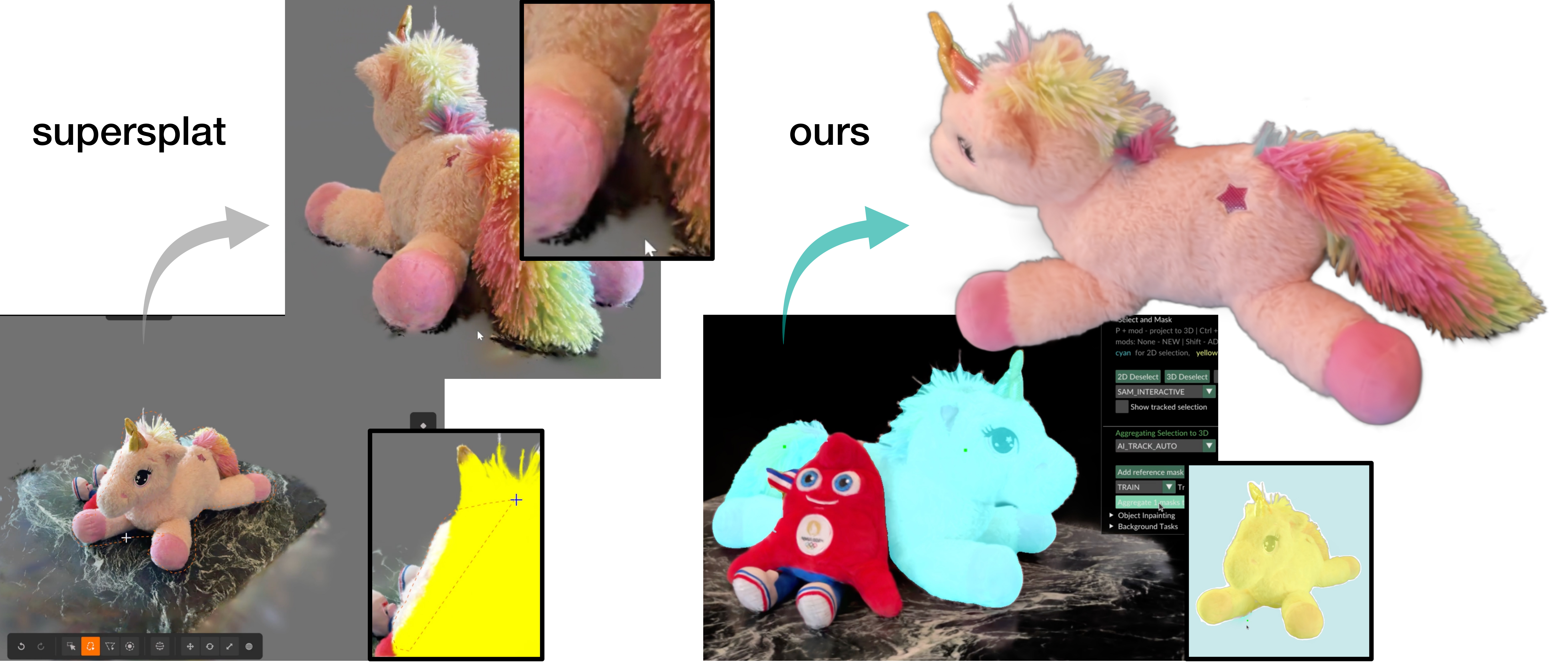}}
        \subfloat[Comparison with GARField.\label{fig:seg_gallery:garfield}]{%
		\includegraphics[width=0.46\linewidth]{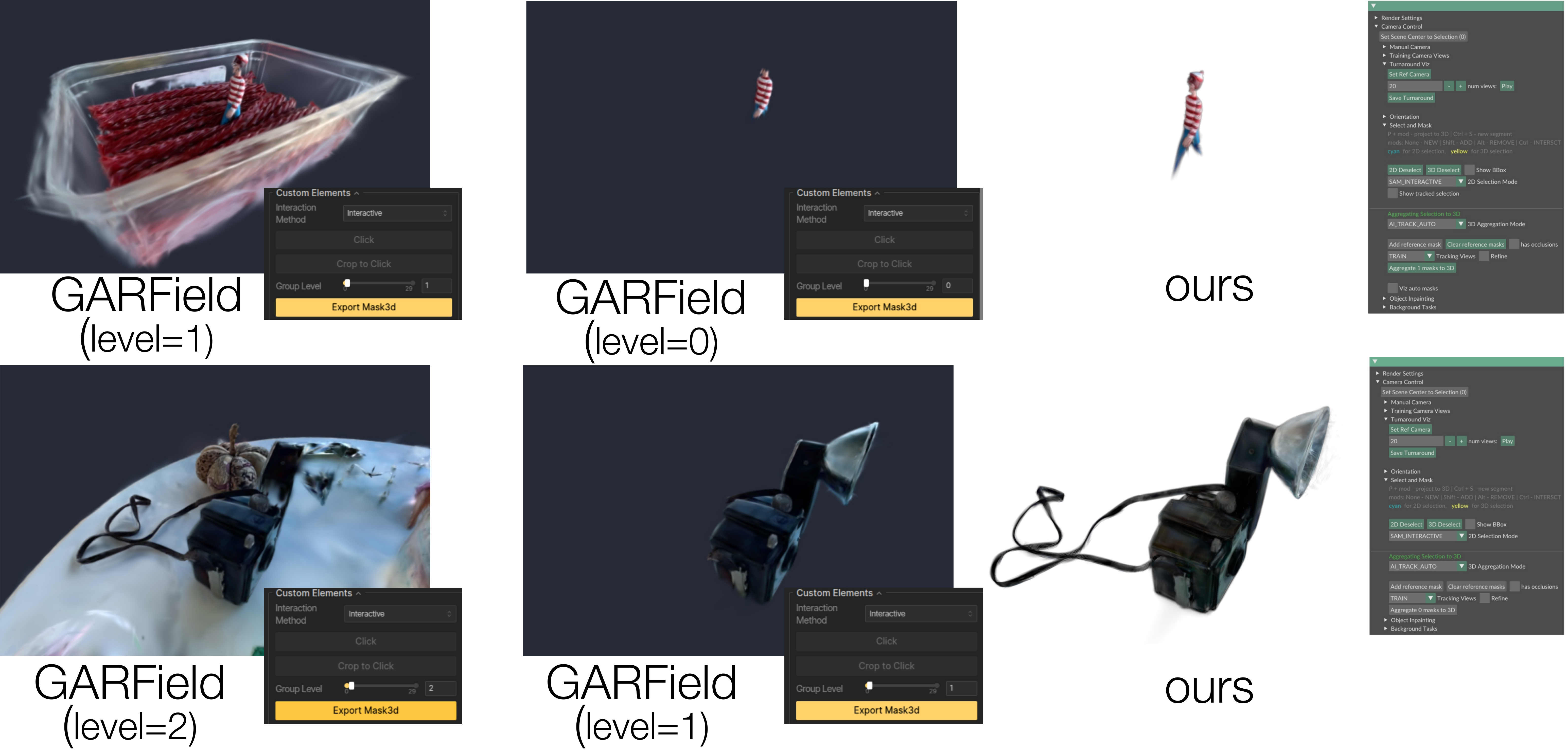}}\\
        \subfloat[Automatic segmentation on LERF.\label{fig:seg_gallery:lerf_auto}]{%
		\includegraphics[width=0.98\linewidth]{img/gallery_seg_lerf.pdf}}\\
	\caption{\textbf{Segmentation Results}: We show the flexibility of our proposed segmentation toolkit (\S\ref{sec:segment}) in the figure above. }
	\label{fig:gallery:compare}
\end{figure*}

Qualitatively, we compare our method against one pre-training method \cite{kim2024garfield} and the training-free GaussianEditor \cite{chen2024gaussianeditor}
approach. We use in-the-wild captured scenes or challenging 360-degree scenes from LERF \cite{lerf2023} that are closer to our target use case.
While in the paper, GaussianEditor reports better performance, we found its segmentation running in around 40s on a higher-end GPU, making any sort of interaction or iteration difficult. Unlike GaussianEditor in practice, our method runs in 1-5 seconds on the same hardware, enabling interactive iteration over the target segmentation and user inputs. Critically, we also allow users to correct both automatic masks or perform segmentation manually (\S\ref{ssec:seg_manual}). We also observe more frequent mistakes for similar examples in Fig.\ref{fig:seg_gallery:gseditor}.
As described in the related work, Garfield \cite{kim2024garfield} suffers from features that are pre-baked during training and can make finer-grained segmentation difficult. On the LERF figurines scene, Garfiled shows similar performance on simple objects, but for more complex examples like "Charlie" in the Twizzlers box or the camera with the strap, Garfield's discrete Group Level parameter 
is insufficient to select the target object and its thin parts (Fig.\ref{fig:seg_gallery:garfield}), while our automatic tracking with a single reference 
easily selects these. We also made an effort to run FlashSplat \cite{shen2025flashsplat}, but ran into technical issues not addressed by the authors (\href{https://github.com/florinshen/FlashSplat/issues/3}{URL}).
Comparing to manual segmentation with software tool SuperSplat \cite{supersplat} (Fig.\ref{fig:seg_gallery:supersplat}), we found SuperSplat very difficult and time-consuming to use, especially when separating objects from surfaces and dealing with stacked objects. On the other hand, our method is a lot faster and user-friendly, providing a good initial segmentation using SAM selection, and allowing additional manual fine-tuning only if necessary. We believe that it strikes just the right
balance between automation and user intervention, allowing selecting virtually any desirable subset of Gaussians given user intent. See \S\ref{sec:applications} for possible applications of such segmentation, and Supplemental Video (soon to be uploaded) for additional qualitative results.

%We NVOS results reported by
%\cite{jain2024gaussiancut} \cite{ying2024omniseg3d} and \cite{shen2025flashsplat}. For qualitative, we could only run Garfield\cite{kim2024garfield} however we couldn't apply our annotation on it as Garfield camera convention are based on Nerfstudio\cite{nerfstudio}. 

% We couldn't run more recent baseline FlashSplat\cite{shen2025flashsplat} as it's missing key pieces to install and run it. We ignored supersplat \cite{supersplat} as the segmentation is only manual and hard to correct\todo{masha correct}

%\subsubsection{Quantitative Results}\label{sec:quant}
%In \ref{tb:seg_eval} we show that our method without pre-segmentation produce 3D mask of better IoU and Accuracy than prior work. While pre-segmentation has lower metrics we show in \todo{masha: I need ref to tab I made that show per-scene metrics} that only "horns\_left" is affected by the change, as the object is cropped out of the reference frame (See \ref{fig:datasets:lerf_mask}), which is not a good representation of a real world usage.

\subsection{Ablations}\label{sec:ablations}
For ablations, we used LERF \cite{lerf2023} figurines scene, a challenging 360-degree scene. 
Because the automatic mask annotations in LERF-Mask \cite{ye2025gaussiangrouping}  are very noisy (Fig.\ref{fig:datasets:lerf_mask}), we instead hand annotated the pretrained 3DGS scene without using any automatic segmentation, using only manual 2D bounding box selections with the frustum projection described in (\S\ref{ssec:seg_manual}). For each object, we additionally annotate input SAM queries and the corresponding mask for one training view (input view: train) and for one additional view outside of the training view trajectory (input view: user), to test different settings of our algorithm. We call this dataset Figurines3DSeg. Unlike quantitative results, computing 2D mask metrics for Acc and mIoU, we use metrics computed over actual Gaussians selected with our method agains ground truth hand annotation. 

\begin{figure}[t!]
	\centering
    \includegraphics[width=0.23\linewidth]{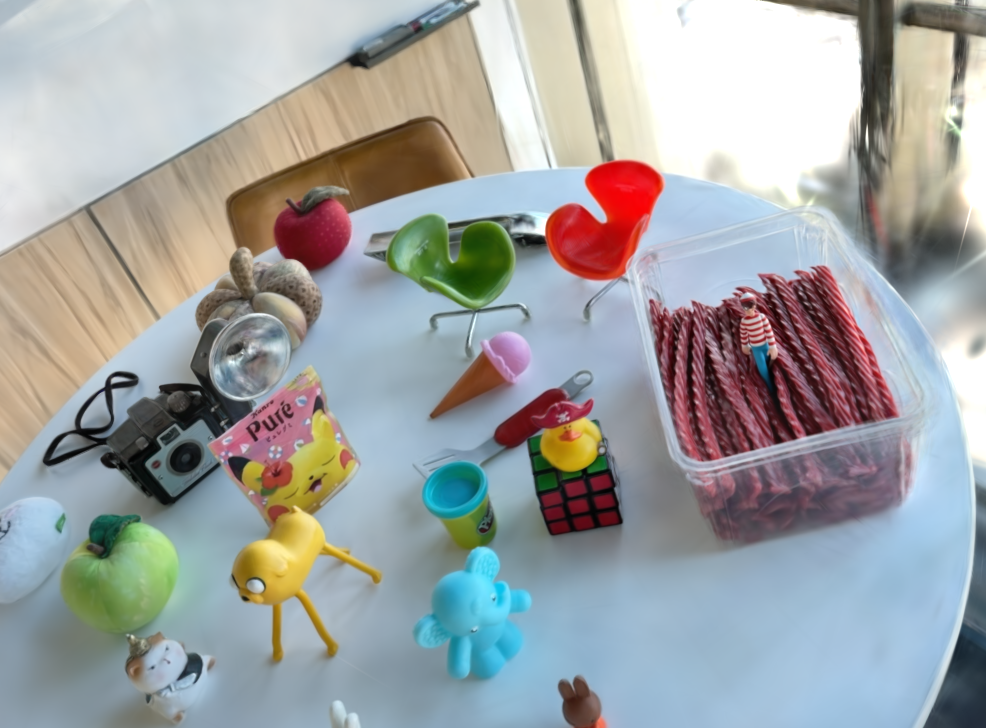}
	\includegraphics[width=0.23\linewidth]{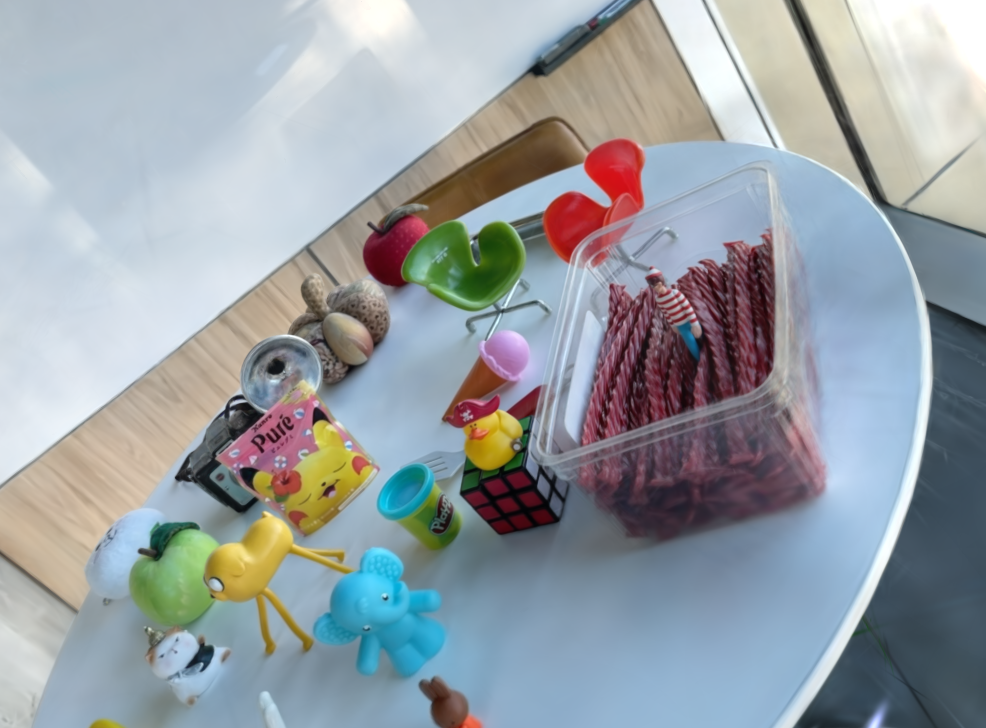}
	\includegraphics[width=0.23\linewidth]{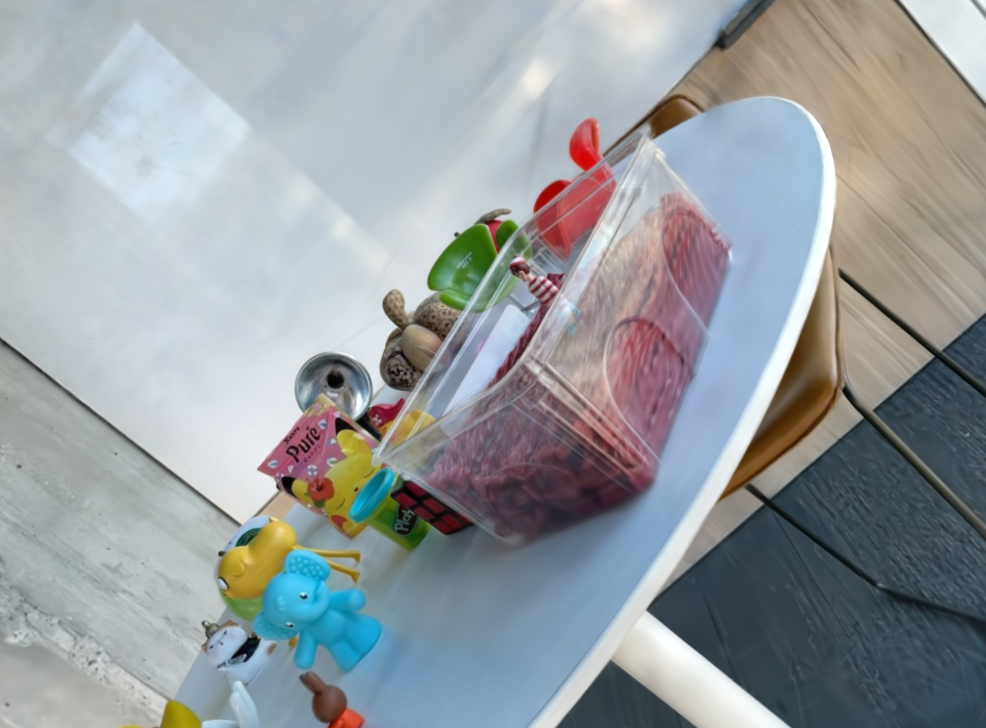}
	\includegraphics[width=0.23\linewidth]{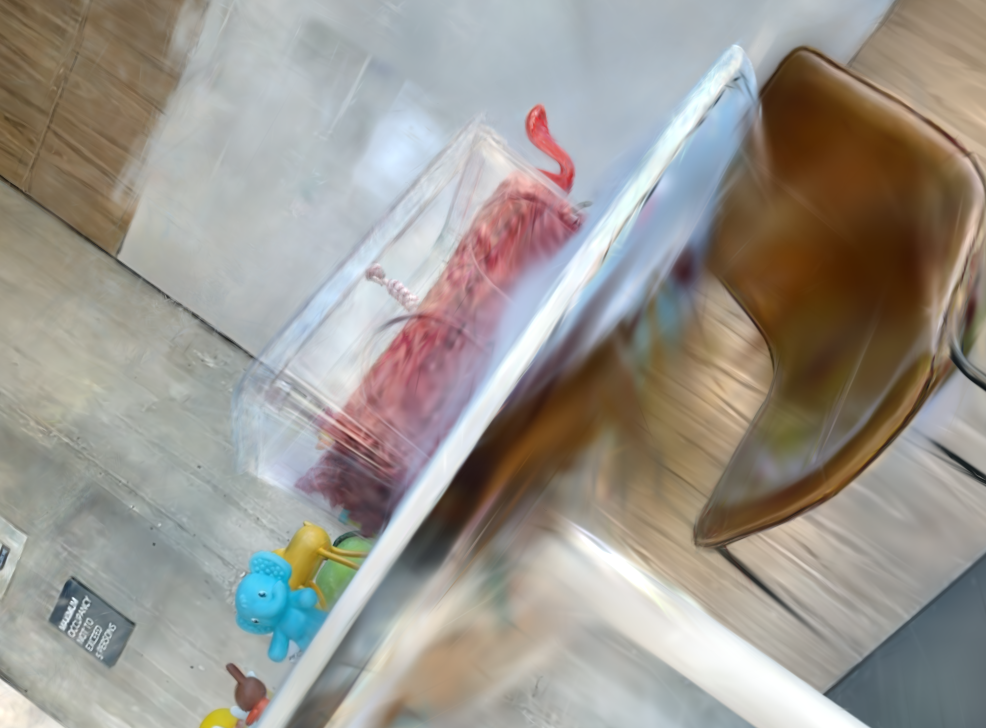}\\
	\includegraphics[width=0.23\linewidth]{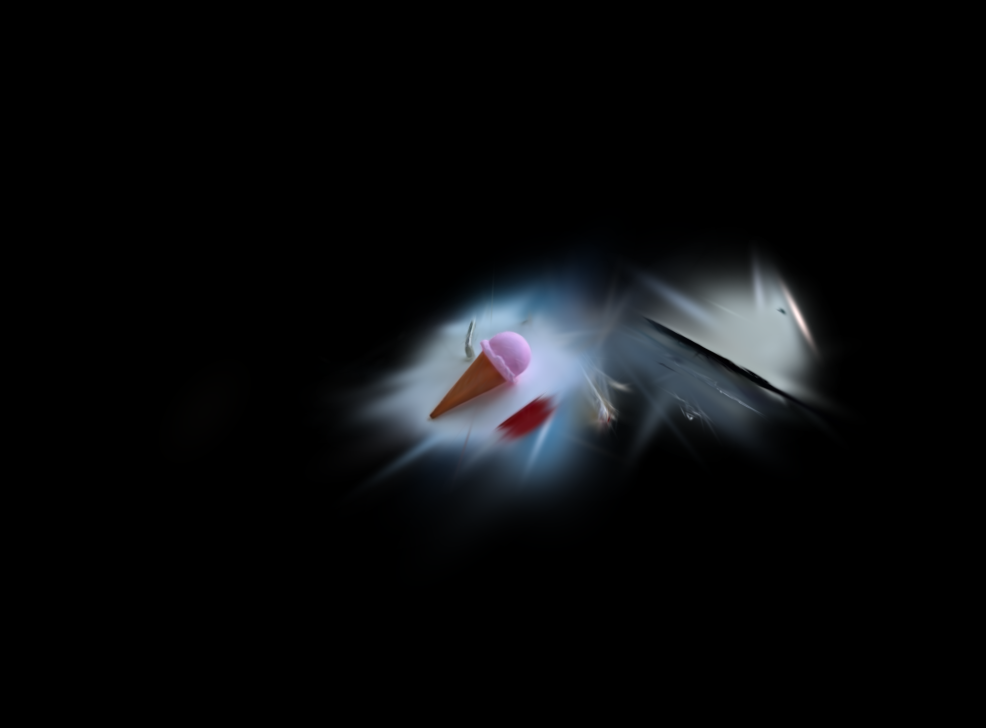}
	\includegraphics[width=0.23\linewidth]{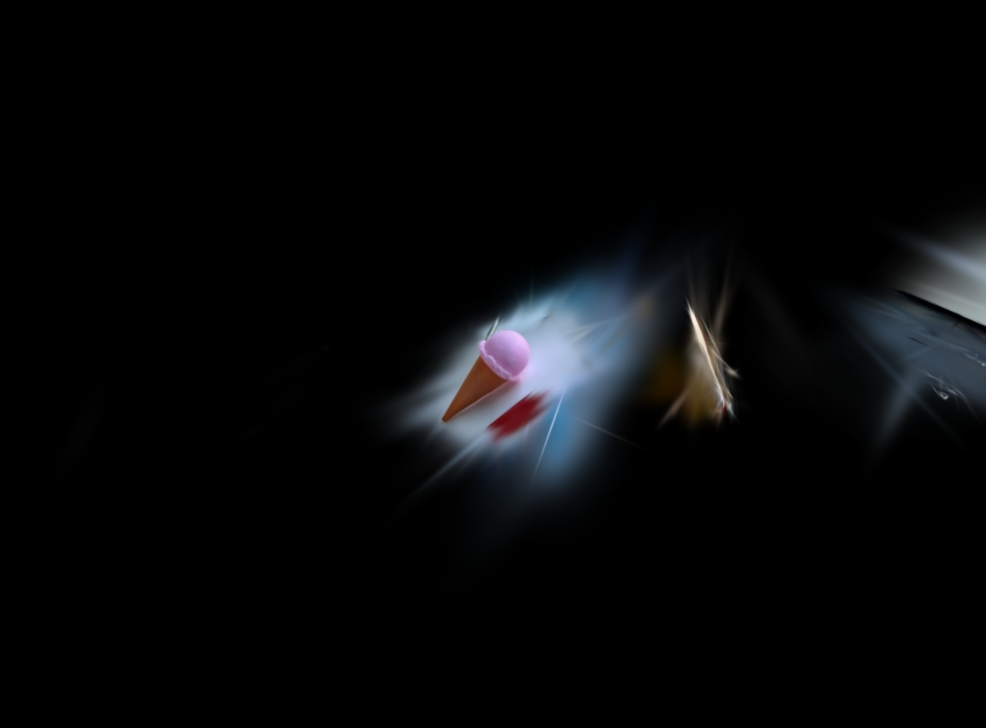}
	\includegraphics[width=0.23\linewidth]{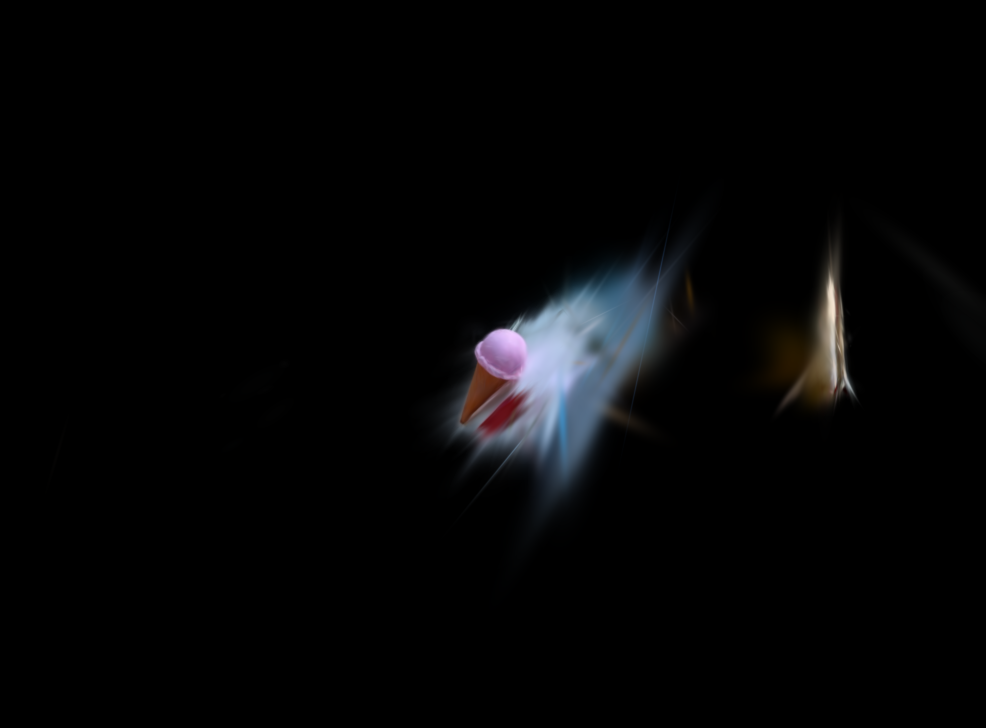}
	\includegraphics[width=0.23\linewidth]{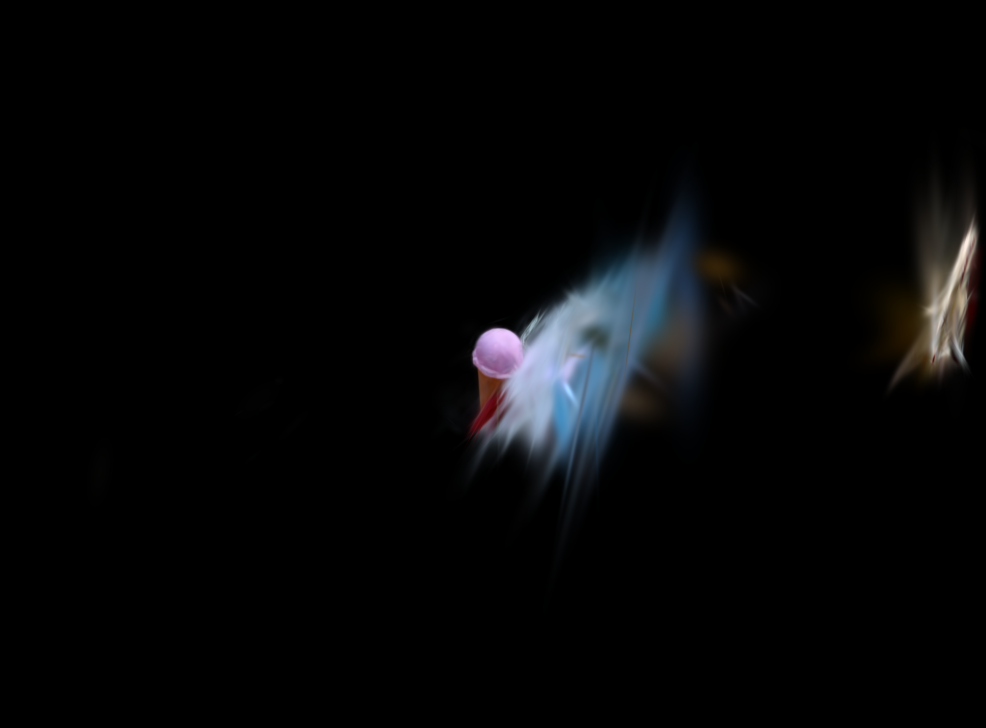}        
	
    \caption{Impact of presegmentation on the inputs to the tracker. Top: Input without presegmentation. Bottom: Input with presegmentation.}
	\label{fig:presegment}
\end{figure}

Because our method tracks masks across viewes (\S\ref{sec:segment}), the choice of views is important. We compare using original scene training views against automatic dense views (\S\ref{ssec:seg_auto:views}) and the impact of the number of views $m$ on quality and speed of our method in  Tb\ref{tb:seg_ablations}. We found $m=50$ to be optimal for performance as well as speed, completing full tracking (including Cutie inference) and 3D segmentation in only 1.5-2.5s. While this is not real-time, this delay is small enough to allow users to iterate with the algorithm in practice. For completeness, we also test with two different kinds of view annotations, showing that our method generalizes to user-selected annotated views, making the interaction a lot less constrained. While we see some degradation when user-selected views are used, the usability still makes this feature attractive. In addition, we see a degradation in both speed and quality of the output when the pre-segmentation (\S\ref{ssec:seg_auto:opt} is turned off. This is because in the scenes with many objects, the tracked views may have occlusions. Pre-segmentation effectively removes some of the occluders in these views (See Fig.\ref{fig:presegment}).

%and the impact of views subsampling and pre-segmentation on the metrics and segmentation speed.
%We show that our best results are with the training views subsampled at 50 views and the strong benefit of pre-segmentation to both quality and speed of segmentation, while subsampling may reduce the segmentation accuracy the method still can be quite robust even at only 10 images.
%We also shows that the turnaround views sampling results stays of good quality and so should be preferred for very large scale scenes or when using a scene without provided cameras.

% \subsubsection{}
% \subsubsection{NVOS}
% We used NVOS\cite{ren-cvpr2022-nvos} for evaluation on LLFF, \cite{cen2023saga} sampled points queries from the scribbles but we found out that the scribbles are incompletes (see \ref{fig:nvos_issue}), and the sampling is oversampled and more importantly not interactive. We fixed that by doing our own sampling interactively, we show on \ref{fig:nvos_custom_sampling} our sampling process and matching masks from SAM\cite{kirillov2023segment}, and further show all the samples matching NVOS inputs in \ref{fig:all_nvos_custom_sampling}.
\begin{table}[t!]
\centering
\begingroup
\small{
\setlength{\tabcolsep}{2pt} % Default value: 6pt
\begin{tabular}{@{}|l||l|l|l|l|l|@{}}
\Xhline{2.0pt}
& \multicolumn{4}{c|}{\textbf{input view}} & \\ \cline{2-5}
& \multicolumn{2}{c|}{train} & \multicolumn{2}{|c|}{user} & \\ \cline{2-5}
 & \multicolumn{1}{c|}{\textbf{mIoU}{$\uparrow$}}  &
 \multicolumn{1}{c|}{\textbf{Acc}{$\uparrow$}} &
 \multicolumn{1}{c|}{\textbf{mIoU}{$\uparrow$}} & 
 \multicolumn{1}{|c|}{\textbf{Acc}{$\uparrow$}} &
 
 \textbf{Speed (s)}\\
\Xhline{2.0pt}
\textbf{type of views (num.)} & & & & & \\
training (all) & 94.0 & 99.2 & 88.9 & 98.9 & 9-12s\\
training (100) & 93.9 & 99.2 & 93.3 & 99.1 & 3-4s\\
training (50) & 93.9 & 99.1 & 94.3 & 99.2 & 1.5-2.5s\\
% training (30) & 93.6 & 99.1 & 93.9 & 99.1 & 1-1.5s\\
training (20) & 92.8 & 99.0 & 94.0 & 99.2 & 0.6-1.2s\\
training (10) & 89.1 & 98.6 & 93.3 & 99.0 & 0.3-1s\\
auto turnaround (100) & 93.8 & 98.9 & 93.2 & 99.0 & 3-4.5s\\
auto turnaround (50) & 93.5 & 98.9 & 93.3 & 99.0 & 1.5-2.5s\\
% auto turnaround (30) & 91.7 & 98.6 & 93.0 & 99.0 & 1s-1.5s\\
auto turnaround (20) & 90.6 & 98.5 & 93.3 & 99.0 & 0.6-1.2s\\
auto turnaround (10) & 89.3 & 98.0 & 92.1 & 98.8 & 0.3-1s\\
\hline
\textbf{no pre-segmentation} & & & & & \\
training (all) & 83.88 & 98.4 & 76.9 & 97.8 & 26-27s\\
auto turnaround (100) & 89.0 & 98.2 & 90.7 & 98.6 & 8-9s\\
\Xhline{2.0pt}
\end{tabular}
}
\endgroup
\caption{Segmentation Ablations on hand-labeled Figurines 3DGS scene.}
\label{tb:seg_ablations}

\end{table}

    \section{Applications}\label{sec:applications}

Our selection and segmentation toolkit feeds into many downstream applications for 3DGS.

\subsection{Interactive Orientation}\label{sec:orient}

\begin{figure*}[h!tbp]
	\centering
        \subfloat[Orientation.\label{fig:seg_gallery:orient}]{%
		\includegraphics[width=0.61\linewidth]{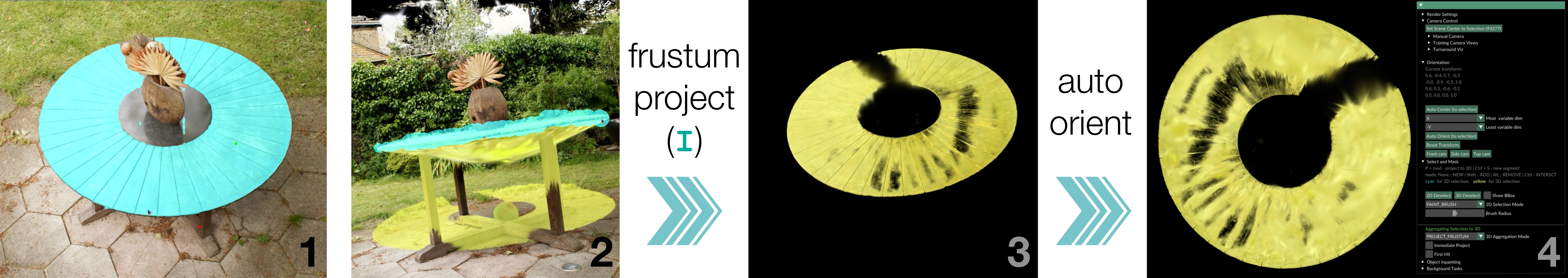}} \\
        \subfloat[Physics simulation of a 3DGS scene segmented with our method, using Simplicits \cite{modi2024simplicits}.\label{fig:app_gallery:simulate}]{%
		\includegraphics[width=0.98\linewidth]{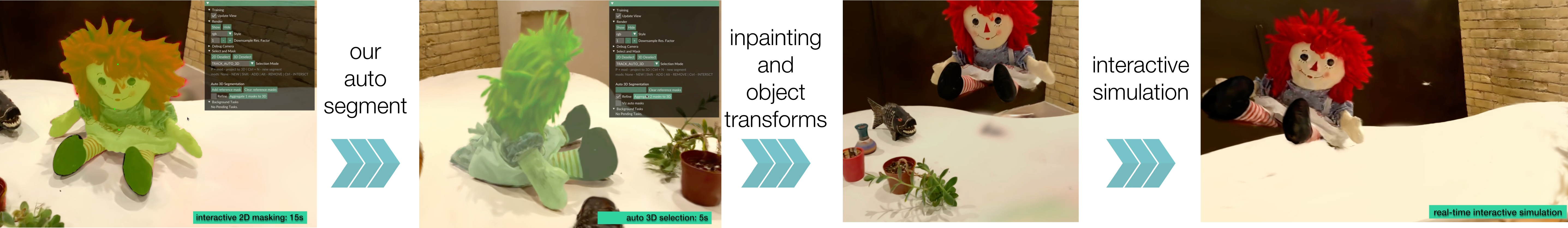}}\\
	\subfloat[Targeted object editing with our method, showing original view (A), selected segment (B), and edited results (C, D).\label{fig:app_gallery:edit0}]{%
		\includegraphics[width=0.98\linewidth]{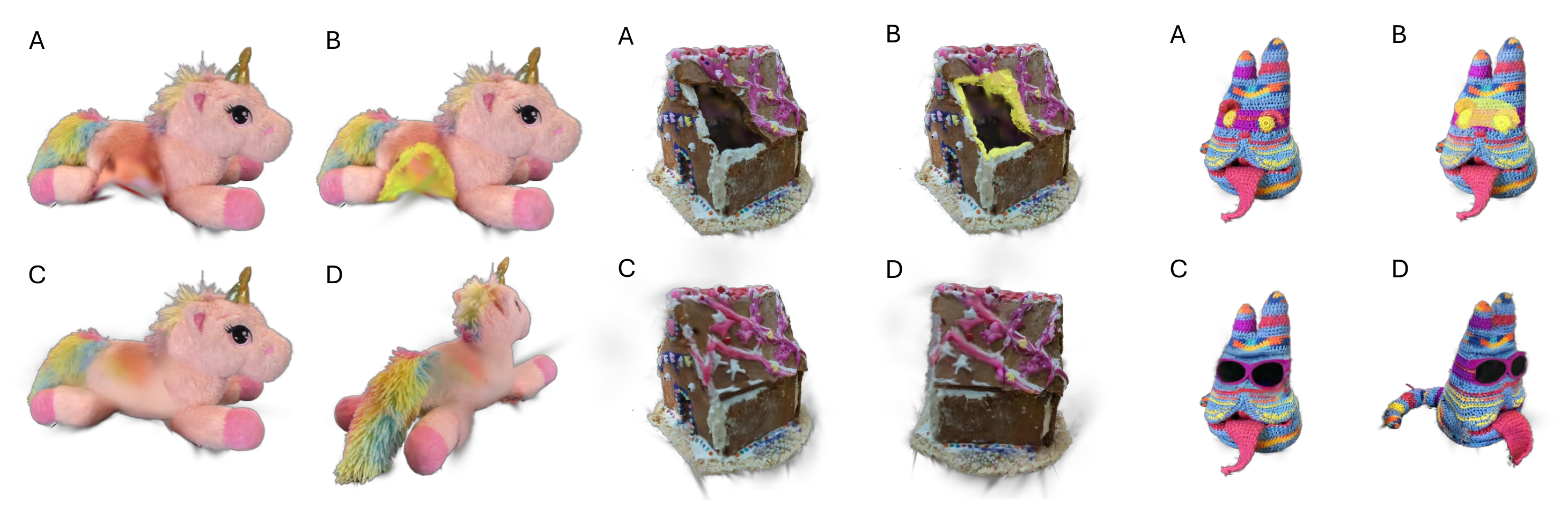}}\\
	\caption{\textbf{Applications}: Our flexible segmentation toolkit can be used to enable user-guided orientation of 3DGS scenes (\S\ref{sec:orient}, Fig.\ref{fig:seg_gallery:orient}). Segmentation and orientation facilitate physics simulation directly over the captured 3DGS scene (\S\ref{ssec:simplicits}, Fig.\ref{fig:app_gallery:simulate}). Controllable selection also enables targeted object editing (\S\ref{sec:refine}, Fig.\ref{fig:app_gallery:edit0}). }
	\label{fig:gallery:app}
\end{figure*}

Orienting a scene is important for consistent camera manipulation, and
is critical for physics simulation, now possible
directly on the 3DGS \cite{modi2024simplicits}. 
We observed that in most cases,
the desirable XYZ axes coincide with the principal directions of variation for 
objects or surfaces. Our auto-orientation module allows users
to automatically align the PCA basis computed over the means of the selected Gaussians with a chosen permutation of world axes.
For example, the user could apply frustum projection to select a flat surface (Fig.\ref{fig:seg_gallery:orient}1-3),
then align Z axis with the axis of least variation (resulting in top capera view in Fig.\ref{fig:seg_gallery:orient} 4) to achieve a consistent gravity direction.
Manually orienting a scene is challenging, and we found this technique simple and effective when used in conjunction with our selection tools (\S\ref{sec:segment}).

\subsection{User-Guided Editing}\label{sec:refine}

Unlike much of generative AI research, our objective is not to hallucinate large 
unseen areas. Instead, our method empowers future directions for targeted editing.
This targeting is enabled through the various selection modes of our tool, providing different levels of precision.
E.g.\ to indicate missing areas occluded parts during capture, the user can select the Gaussians around the hole using the paint tool. This selection could be
used to generate masks for AI inpainting and be used to add new Gaussians. 
For editing existing areas, SAM selection can be used to efficiently remove parts. The paint tool can also be used to draw on interesting shapes to inpaint. 

%This way, the selected Gaussians could be used directly to create more precise masks for inpainting.

As a prototype towards targeted local editing, we experiment with a Video inpainting network that can be used to inpaint and optimize 
Gaussians only in the user-selected regions. We build our method on CogVideoX's Image-to-Video model \cite{yang2024cogvideox} as a strong video generation foundation and insert CamCo's epipolar attention module \cite{xu2024camco} for camera control and 3D consistency. The inputs to our video model are masked video frames, binary masks for each frame, plucker embeddings and epipolar lines from camera parameters. 
The first frame of the video is a reference image that is generated by Stable Diffusion XL inpainting model. 
The typical user workflow would be to inpaint the first view using a text prompt, once satisfied, apply our video model to generate additional views starting from the first view. 
To propagate these changes to 3D, we remove the Gaussians that are selected and initialize new gaussians randomly inside the selection bounding box then train using the inpainted views. %During optimization, we prune any Gaussians that is added outside the selection box to ensure no floaters are added. 
We show these early experiments to demonstrate how precise selection can help guide editing. Because we do not train on the unselected original Gaussians, we do not modify the look of the unselected area. We are excited to explore this application in future research.

\begin{figure}[t!]
	\centering
	
		\includegraphics[width=0.95\linewidth]{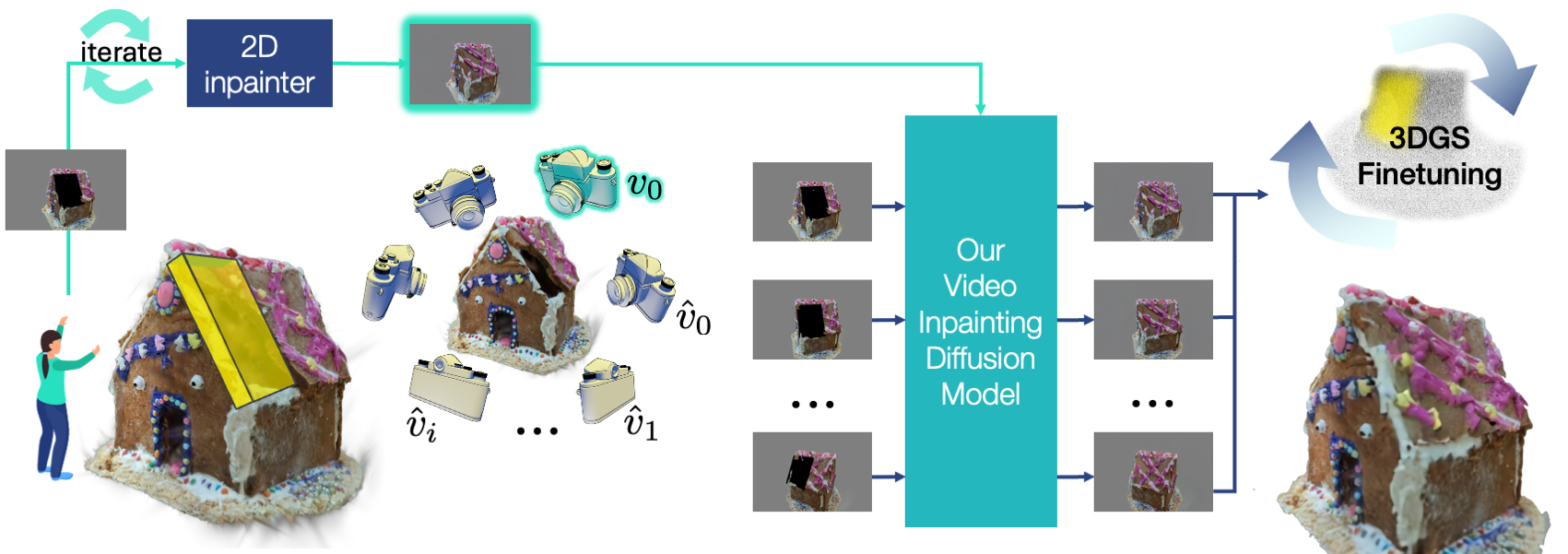}
       
	\caption{\textbf{Object completion}(\S\ref{sec:refine}): after the user marks region to be completed and iterates on a 2D inpainting for one view, we apply a custom video inpainting model to propagate the result to multi-view consistent frames, which are used to fine-tune the 3DGS object.}% (a); our video diffusion block is shown in b.}
	\label{fig:editing}
\end{figure}

\subsection{Simulating Splat Objects}\label{ssec:simplicits}

With the ability to segment, orient (\S\ref{sec:orient}) and refine (\S\ref{sec:refine}) it is becoming
increasingly feasible to convert raw in-the-wild captures to simulated scenes. For example, 
3DGS objects can be directly simulated using Simplicits \cite{modi2024simplicits} (See Fig.\ref{fig:app_gallery:simulate} and video). Our flexible selection and segmentation toolkit also
makes it possible for users to select individual object parts and assign different materials, for
example to make the hair of the doll more deformable. This application is now directly possible
using existing simulation techniques and our user-guded selection.

\section{Conclusion}

In conclusion, we presented \ourmodel{}, a suite of versatile interactive selection and segmentation
tools for 3D Gaussian Splats with AI and user in the loop. We believe that a flexible solution to
user guided segmentation is a necessary foothold for many applications.

% \bibliographystyle{ACM-Reference-Format}
% \bibliography{main}
    \putbib[main]
\end{bibunit}

\newpage


\begin{acks}
\end{acks}

\clearpage

\appendix
\begin{bibunit}[ACM-Reference-Format]
    \section{Comparison to Standard Software}
Under controlled object-centric conditions (even as simple as suspending an object on a string), it is possible to capture data where one can easily select objects using simple techniques like bounding boxes. However, there are many scenarios 'in the wild' where this might not be possible.  Editing task, like moving or removing specific objects, are heavily dependent on good selection tools. A number of standard software applications address this problem:
Gaussian splat capturing solutions like \textsf{Scaniverse} \cite{scaniverse} or \textsf{Polycam} \cite{polycam} allow bounding volume based selection. In addition \textsf{KIRI Engine} \cite{kiriengine} as well as \textsf{Postshot} \cite{postshot} feature brush based selection, in many cases the focus here is object extraction via deletion of unwanted parts of the scan. 
\textsf{Gaussian Splatting} \cite{gaussianae, gaussiannuke} for \textsf{After Effects} \cite{aftereffects} and \textsf{Nuke} \cite{nuke} allow integration of Gaussian Splats into 2.5d image compositing pipelines - their feature set matches the above.
Dedicated Gaussian splat web apps like \textsf{Supersplat} \cite{Supersplat} as well as extensions to game engines \textsf{XVERSE 3D-GS UE Plugin} \cite{xverse3dgs} for \textsf{Unreal Engine} \cite{unreal} or \textsf{SplatVFX} \cite{splatvfx} for \textsf{Unity} \cite{unity} expand the editing capabilities by allowing manipulation using their standard tooling on a point level. 
Extensions to standard 3D animation software like \textsf{NerfStudio} \cite{nerfstudioblender} for \textsf{Blender} \cite{Blender} offer a similar feature set. In addition to cleanup, segmentation is used here often to extract or modify individual features from a scan for downstream use.
\textsf{GSOPs} \cite{gsops} for \textsf{Houdini} \cite{houdini} not only opens up the the use Houdini's direct and procedural tools for direct and indirect feature selection for Gaussian Splat models, but also adds automatic clustering and segmentation based on the DBSCAN algorithm \cite{dbscan}.

While user-guided feature selection in Gaussian Splatting data is the main focus of this paper, we nevertheless consider feature selection in point cloud or volumetric data for industrial or medical uses closely related: \textsf{ReCap Pro} \cite{recap}, \textsf{Reality Capture} \cite{realitycapture} and \textsf{PIX4Dmatic} \cite{pix4dmatic} are photogrammetry solutions used in surveying, where a user might need to identify elements of a scan (for example part of a worksite) for further inspection. 

In addition to the standard selection tools found in 3D-animation software, we can find automatic and semi-automatic solutions that help dealing with often large data volumes: \textsf{Segments AI} \cite{segmentsai}, \textsf{Pointly} \cite{pointly}, \textsf{Metashape} \cite{metashape} and \textsf{ArcGIS} \cite{arcgis} offer automatic machine learning based feature classification for large scale data-labeling needs with application in fields like city mapping or autonomous driving.  
In medical imaging, a radiologist might need to select a organ from a volumetric CT scan for further analysis. \textsf{3D Slicer} \cite{slicer} is a popular tool that allows to automatically select such features from a volumetric scan using machine learning with extensions like \textsf{TotalSegmentator} \cite{totalsegmentator}. Here, 'region growing' is a semi-automatic technique  where a user identifies 'seed points' in features used as reference data points for similarity based selection expansion.

With Gaussian Splatting being adopted for industry use-cases, we believe our research to be applicable and complementary to existing tooling in these domains, too.

\section{Evaluation Details}

Breakdown of the NVOS results per testcases is presented in Table 1, showing that pre-segmentation only affects one example.

\begin{table}[t!]
\centering

\begingroup
%\footnotesize{
\setlength{\tabcolsep}{2pt} % Default value: 6pt
\begin{tabular}{@{}|l||ll|ll|@{}}
\Xhline{2.0pt}
& \multicolumn{2}{c|}{Ours} & \multicolumn{2}{c|}{Ours (without pre-segmentation)} \\
& \textbf{mIoU} & \textbf{Acc} & \textbf{mIoU} & \textbf{Acc} \\
\Xhline{2.0pt}
fern & 83.6 & 94.7 & 83.6 & 94.8\\
flower & 97.9 & 99.5 & 97.9 & 99.5\\
fortress & 98.5 & 99.7 & 98.3 & 99.7\\
horns left & 0.0 & 94.0 & 92.7 & 99.4\\
horns center & 97.4 & 99.4 & 97.4 & 99.4\\
leaves & 96.4 & 99.8 & 96.6 & 99.8\\
orchids & 96.6 & 99.2 & 96.7 & 99.2\\
trex & 89.2 & 98.5 & 89.2 & 98.6\\
\Xhline{2.0pt}
\end{tabular}
%}
\endgroup
\caption{NVOS Segmentation evaluation. The ``horns\_left'' is failing with pre-segmentation because the object is partially out of frame in the input mask.}
\label{tb:nvos_eval}

\end{table}

    \putbib[software]
\end{bibunit}

\end{document}

\title{{\ourmodel}: Interactive Tools for Gaussian Splat Selection
\\with AI and Human in the Loop}

%AI-Assisted Segmentation and Completion
%of Gaussian Splat Objects with Human-in-the-Loop}

% Interactive Tools for Gaussian Splat Selection
% and Editing with AI and Human in the Loop

% CraftGS: Interactive AI-powered Segmentation and Refinement for Crafting High-Quality Gaussian Splat Objects from In-the-Wild Reconstructions
% 
% User and AI Assisted Distillation of Interactive 3D Objects
% from 3D Gaussian Splat Reconstrucitons
%
% Verbs: crafting, curating, fashioning
% Segmentation, Object Completion, Refinement
%
% Pithy Tagline: Description
\author{Ben Trovato}
\authornote{Both authors contributed equally to this research.}
\email{trovato@corporation.com}
\orcid{1234-5678-9012}
\author{G.K.M. Tobin}
\authornotemark[1]
\email{webmaster@marysville-ohio.com}
\affiliation{%
  \institution{Institute for Clarity in Documentation}
  \city{Dublin}
  \state{Ohio}
  \country{USA}
}

\author{Lars Th{\o}rv{\"a}ld}
\affiliation{%
  \institution{The Th{\o}rv{\"a}ld Group}
  \city{Hekla}
  \country{Iceland}}
\email{larst@affiliation.org}

\author{Valerie B\'eranger}
\affiliation{%
  \institution{Inria Paris-Rocquencourt}
  \city{Rocquencourt}
  \country{France}
}

\author{Aparna Patel}
\affiliation{%
 \institution{Rajiv Gandhi University}
 \city{Doimukh}
 \state{Arunachal Pradesh}
 \country{India}}

\author{Huifen Chan}
\affiliation{%
  \institution{Tsinghua University}
  \city{Haidian Qu}
  \state{Beijing Shi}
  \country{China}}

\author{Charles Palmer}
\affiliation{%
  \institution{Palmer Research Laboratories}
  \city{San Antonio}
  \state{Texas}
  \country{USA}}
\email{cpalmer@prl.com}

\author{John Smith}
\affiliation{%
  \institution{The Th{\o}rv{\"a}ld Group}
  \city{Hekla}
  \country{Iceland}}
\email{jsmith@affiliation.org}

\author{Julius P. Kumquat}
\affiliation{%
  \institution{The Kumquat Consortium}
  \city{New York}
  \country{USA}}
\email{jpkumquat@consortium.net}

\maketitle

\begin{bibunit}{ACM-Reference-Format}
    \section{Comparison to Standard Software}
Under controlled object-centric conditions (even as simple as suspending an object on a string), it is possible to capture data where one can easily select objects using simple techniques like bounding boxes. However, there are many scenarios 'in the wild' where this might not be possible.  Editing task, like moving or removing specific objects, are heavily dependent on good selection tools. A number of standard software applications address this problem:
Gaussian splat capturing solutions like \textsf{Scaniverse} \cite{scaniverse} or \textsf{Polycam} \cite{polycam} allow bounding volume based selection. In addition \textsf{KIRI Engine} \cite{kiriengine} as well as \textsf{Postshot} \cite{postshot} feature brush based selection, in many cases the focus here is object extraction via deletion of unwanted parts of the scan. 
\textsf{Gaussian Splatting} \cite{gaussianae, gaussiannuke} for \textsf{After Effects} \cite{aftereffects} and \textsf{Nuke} \cite{nuke} allow integration of Gaussian Splats into 2.5d image compositing pipelines - their feature set matches the above.
Dedicated Gaussian splat web apps like \textsf{Supersplat} \cite{Supersplat} as well as extensions to game engines \textsf{XVERSE 3D-GS UE Plugin} \cite{xverse3dgs} for \textsf{Unreal Engine} \cite{unreal} or \textsf{SplatVFX} \cite{splatvfx} for \textsf{Unity} \cite{unity} expand the editing capabilities by allowing manipulation using their standard tooling on a point level. 
Extensions to standard 3D animation software like \textsf{NerfStudio} \cite{nerfstudioblender} for \textsf{Blender} \cite{Blender} offer a similar feature set. In addition to cleanup, segmentation is used here often to extract or modify individual features from a scan for downstream use.
\textsf{GSOPs} \cite{gsops} for \textsf{Houdini} \cite{houdini} not only opens up the the use Houdini's direct and procedural tools for direct and indirect feature selection for Gaussian Splat models, but also adds automatic clustering and segmentation based on the DBSCAN algorithm \cite{dbscan}.

While user-guided feature selection in Gaussian Splatting data is the main focus of this paper, we nevertheless consider feature selection in point cloud or volumetric data for industrial or medical uses closely related: \textsf{ReCap Pro} \cite{recap}, \textsf{Reality Capture} \cite{realitycapture} and \textsf{PIX4Dmatic} \cite{pix4dmatic} are photogrammetry solutions used in surveying, where a user might need to identify elements of a scan (for example part of a worksite) for further inspection. 

In addition to the standard selection tools found in 3D-animation software, we can find automatic and semi-automatic solutions that help dealing with often large data volumes: \textsf{Segments AI} \cite{segmentsai}, \textsf{Pointly} \cite{pointly}, \textsf{Metashape} \cite{metashape} and \textsf{ArcGIS} \cite{arcgis} offer automatic machine learning based feature classification for large scale data-labeling needs with application in fields like city mapping or autonomous driving.  
In medical imaging, a radiologist might need to select a organ from a volumetric CT scan for further analysis. \textsf{3D Slicer} \cite{slicer} is a popular tool that allows to automatically select such features from a volumetric scan using machine learning with extensions like \textsf{TotalSegmentator} \cite{totalsegmentator}. Here, 'region growing' is a semi-automatic technique  where a user identifies 'seed points' in features used as reference data points for similarity based selection expansion.

With Gaussian Splatting being adopted for industry use-cases, we believe our research to be applicable and complementary to existing tooling in these domains, too.

\section{Evaluation Details}

Breakdown of the NVOS results per testcases is presented in Table 1, showing that pre-segmentation only affects one example.

\begin{table}[t!]
\centering

\begingroup
%\footnotesize{
\setlength{\tabcolsep}{2pt} % Default value: 6pt
\begin{tabular}{@{}|l||ll|ll|@{}}
\Xhline{2.0pt}
& \multicolumn{2}{c|}{Ours} & \multicolumn{2}{c|}{Ours (without pre-segmentation)} \\
& \textbf{mIoU} & \textbf{Acc} & \textbf{mIoU} & \textbf{Acc} \\
\Xhline{2.0pt}
fern & 83.6 & 94.7 & 83.6 & 94.8\\
flower & 97.9 & 99.5 & 97.9 & 99.5\\
fortress & 98.5 & 99.7 & 98.3 & 99.7\\
horns left & 0.0 & 94.0 & 92.7 & 99.4\\
horns center & 97.4 & 99.4 & 97.4 & 99.4\\
leaves & 96.4 & 99.8 & 96.6 & 99.8\\
orchids & 96.6 & 99.2 & 96.7 & 99.2\\
trex & 89.2 & 98.5 & 89.2 & 98.6\\
\Xhline{2.0pt}
\end{tabular}
%}
\endgroup
\caption{NVOS Segmentation evaluation. The ``horns\_left'' is failing with pre-segmentation because the object is partially out of frame in the input mask.}
\label{tb:nvos_eval}

\end{table}

% \bibliographystyle{ACM-Reference-Format}
% \bibliography{software}
    \putbib{software}
\end{bibunit}